\documentclass[10pt,twocolumn,letterpaper]{article}

\usepackage{cvpr}

\definecolor{isabelline}{RGB}{244, 240, 236}
\definecolor{kaiming-green}{RGB}{57,181,74} 
\definecolor{lightblue}{RGB}{240, 246, 254}
\definecolor{gray}{RGB}{242,242,242}

\usepackage[toc,page,header]{appendix}
\usepackage{minitoc}
\doparttoc 
\faketableofcontents 

\usepackage{float}
\usepackage{xspace}
\usepackage{graphicx}
\usepackage{enumitem}
\usepackage{amsmath}
\usepackage{multirow}
\usepackage{pifont}
\usepackage{colortbl}
\usepackage{wrapfig}
\usepackage{subcaption}
\usepackage{bm}
\usepackage{threeparttable}
\usepackage{ragged2e}

\usepackage{algorithmic}
\usepackage{amssymb}

\usepackage{setspace}
\usepackage[ruled,norelsize]{algorithm2e}

\definecolor{commentcolor}{RGB}{110,154,155}   
\newcommand{\PyComment}[1]{\ttfamily\textcolor{commentcolor}{\# #1}}  
\newcommand{\PyCode}[1]{\ttfamily\textcolor{black}{#1}} 

\definecolor{cvprblue}{rgb}{0.21,0.49,0.74}
\usepackage[pagebackref,breaklinks,colorlinks,allcolors=cvprblue]{hyperref}

\makeatother

\title{Revisiting Data Challenges of Computational Pathology: A Pack-based Multiple Instance Learning Training Framework}

\author{%
  Wenhao Tang$^{1}$\thanks{Equal contribution.} \and Heng Fang$^{2}$\footnotemark[1] \and Ge Wu$^{1}$ \and Xiang Li$^{1,3}$\thanks{Corresponding author} \and Ming-Ming Cheng$^{1,3}$\footnotemark[2]
  \and 
  $^{1}$VCIP, School of Computer Science, Nankai University\\
  $^{2}$Huazhong University of Science and Technology\\
  $^{3}$Nankai International Advanced Research Institute (Shenzhen Futian) \\
}

\begin{document}
\maketitle
\begin{abstract}
Computational pathology (CPath) digitizes pathology slides into whole slide images (WSIs), enabling analysis for critical healthcare tasks such as cancer diagnosis and prognosis. 
However, WSIs possess extremely long sequence lengths, significant length variations (from 200 to 200K), and limited supervision. 
These extreme length variations lead to high data heterogeneity and redundancy.
Conventional methods often compromise on training efficiency and optimization to preserve such heterogeneity under limited supervision.
To comprehensively address these challenges, we propose a pack-based MIL framework.
It packs multiple sampled, variable-length feature sequences into fixed-length ones, enabling batched training while preserving data heterogeneity.
Moreover, we introduce a residual branch that composes discarded features from multiple slides into a \textit{hyperslide} which is trained with tailored labels. It offers multi-slide supervision while mitigating feature loss from sampling. 
Meanwhile, an attention-driven downsampler is introduced to compress features in both branches to reduce redundancy.
By alleviating these challenges, our approach achieves an accuracy improvement of up to 8\% while using only 12\% of the training time in the PANDA (UNI). 
Extensive experiments demonstrate that focusing data challenges in CPath holds significant potential in the era of foundation models. 
The code is~\href{https://github.com/FangHeng/PackMIL}{here}.
\end{abstract}

\section{Introduction}
\label{sec:intro}

\begin{figure*}[t]
    \centering
    \includegraphics[width=.9\textwidth]{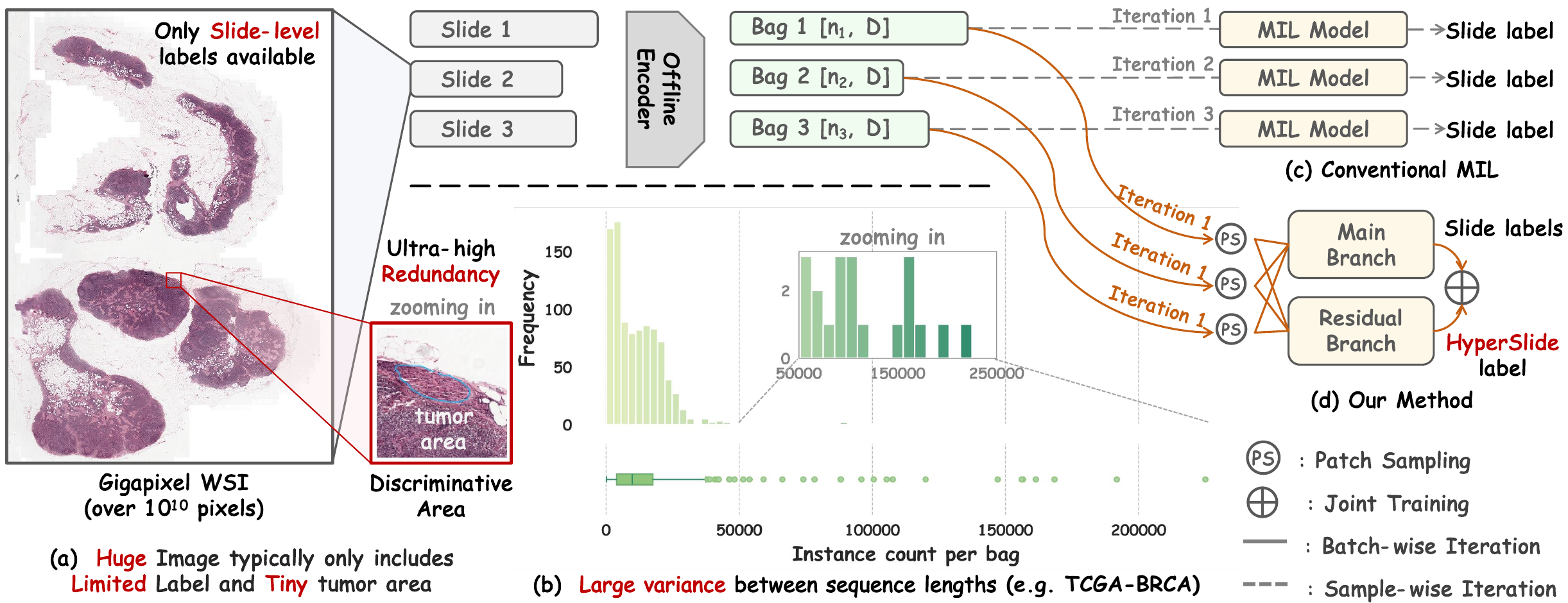}
    \caption{
    \textbf{(a, b)}: 
    WSIs present significant data challenges, including high heterogeneity stemming from highly variable sequence lengths and diverse morphology, massive data redundancy, and limited supervision.
    \textbf{(c)}: Conventional methods train with batchsize of 1 to preserve data heterogeneity, suffering from training inefficiency
    and instability.
    \textbf{(d)}: Our pack-based framework packs variable-length sequences to preserve scale information. It further introduces a residual branch to model inter-slide correlations, constructing a \textit{hyperslide} that retains all morphological features and enrich limited supervision. This approach maintains data heterogeneity while enabling batched training.
    }
    \label{fig:intro}
\end{figure*}

Computational pathology (CPath)~\cite{song2023artificial, cifci2023ai} represents a rapidly evolving interdisciplinary research domain that integrates advanced computer vision techniques and pathology to facilitate accurate and efficient interpretation of histopathological images. Central to CPath are whole slide images (WSIs, slides), digitized pathology slides with gigapixel resolution. It enables comprehensive microscopic analysis to support critical healthcare tasks such as cancer sub-typing~\cite{ilse2018attention,zhang2022dtfd,tu2022dual}, grading~\cite{panda}, and prognosis~\cite{wen2023deep,yao2020whole}. Patching strategies help researchers effectively process these gigapixel images within hardware constraints. However, patches derived from WSIs present following challenges:
as shown in Fig.\ref{fig:intro}, 
\textcolor{magenta}{\textbf{1)}} they possess \textit{extremely long sequence lengths} and \textit{significant sequence length variations} (e.g., from 200 to 200K in TCGA-BRCA-Survival). 
Such extreme distributions in sequence length, 
coupled with \textit{diverse morphological characteristics}, contribute to \textit{data heterogeneity}, which is substantial for
CPath tasks.
\textcolor{magenta}{\textbf{2)}} and introduce input \textit{redundancy challenges} for CPath algorithms.
\textcolor{magenta}{\textbf{3)}} Moreover, due to the gigapixel resolution and specialization, WSIs typically have \textit{only slide-level annotations}, lacking more supervision that matches the complex input.

Current two-stage multiple instance learning (MIL)~\cite{maron1997mil_1} paradigm~\cite{clam} is a compromise resulting from high data heterogeneity and limited supervision.
This paradigm employs a pre-trained encoder to extract offline patch (instance) features, and uses a MIL model to produce slide-level (bag-level) results. 
Due to data challenges, it suffers from training inefficiency and instability.
Specifically, with significant variations in sequence length across slides, mainstream methods typically process data with a batchsize of 1 during training \cite{shao2021transmil,li2024dynamic}. 
While these approaches preserve whole-slide heterogeneity, training with a batchsize of 1 is inefficient (e.g., training TransMIL~\cite{shao2021transmil} on the PANDA~\cite{panda} dataset requires over 50 RTX3090 GPU-hours) and may yield suboptimal performance~\cite{koga2025attention}. 
A few methods~\cite{campanella2019clinical,liu2024attention} attempt to enable batched training by sampling or padding all sequences to a uniform length; however, this approach can lead to a loss of data heterogeneity and important features, especially affecting complex methods and tasks.

To comprehensively address three data challenges, we propose a novel pack-based MIL framework.
Inspired by recent advancements in large model~\cite{pouransari2024dataset,krell2021efficient,dehghani2023patch,wang2024qwen2}, it packs multiple variable-length sequences into a single fixed-length sequence to enable batched training while preserving data heterogeneity.
However, leveraging packing strategies for effective batched training in CPath is far from straightforward.
The excessive length of packed sequences hinders training, necessitating patch sampling, which still leads to feature loss. 
Therefore, we split the input features into main and residual branches, packing the kept and discarded features, respectively, to minimize sampling-induced feature loss.
In the main branch, we employ masks to maintain the independence of different slides within a pack.
Conversely, the residual branch treats discarded features from multiple slides in the same pack as a single \textit{hyperslide}. To train this \textit{hyperslide} effectively, we introduce task-specific hyperslide labels and loss functions.
Crucially, this approach effectively offers multi-slide supervision.

While some outstanding works have explored supplementary supervision~\cite{zhang2022dtfd,shao2023lnpl,brussee2025graph,fang2024sam}, most focus on intra-slide modeling (e.g., instance-level or pseudo-bag), neglecting inter-slide relationships. Pathology slides from the same spatial and tissue origin exhibit consistent morphological characteristics~\cite{lin2025impact,chen2022fast,kaczmarzyk2024open}. Learning inter-slide correlations allows the \textit{hyperslide} to provide the model with a more comprehensive perspective, enabling the discovery of more generalizable pathological features. Furthermore, we propose an attention-driven downsampler to compress features for reducing input redundancy within both branches. To validate our framework, we conducted extensive experiments using features from foundation models. Results demonstrate that our approach consistently improves multiple baselines by effectively mitigating the data challenges inherent in CPath. Specifically, it delivers substantial performance gains (e.g., \textcolor{magenta}{+11\%} accuracy on PANDA) while improving training efficiency (\textcolor{magenta}{$\sim8 \times$} speedup on PANDA). Our contributions are:

\begin{itemize}
    \item We revisit the data challenges in CPath, like high heterogeneity, high redundancy, and limited supervision. 
    Considering these challenges, we propose an efficient and effective pack-based MIL framework that enables reliable training while preserving data heterogeneity.
    \item We construct the \textit{hyperslide} from discarded features during the packing. Corresponding task-specific hyperslide labels and loss functions are designed. This strategy not only minimizes sampling-induced feature loss but also introduces multi-slide supervision. It provides the model with a more comprehensive perspective, thereby improving CPath performance.
    \item We propose an attention-driven downsampler to compress redundant features during the training process. 
    With extensive experiments, we validate the effectiveness of the proposed approach, summarize practical guidelines for batched CPath training, and highlight the significant potential of addressing data challenges in the era of FM.
\end{itemize}

\section{Related Works}
\label{sec:related_works}

\textbf{Supervision in Computational Pathology.}
Recent CPath advancements leverage MIL to reduce annotation burden.
Using only slide labels, MIL has evolved with mechanisms like attention~\cite{ilse2018attention,tang2023mhim,zhang2024attention,zhang2025aem,dong2025fast}, clustering~\cite{lin2023interventional}, Transformers~\cite{shao2021transmil,jaume2024transcriptomics,fourkioti2023camil}, and GNNs~\cite{wang2021hierarchical,eastwood2023mesograph} to improve interpretability and accuracy.
Complementing pure MIL methods, pseudo-labeling strategies have emerged as powerful techniques, encompassing instance-level pseudo-labeling~\cite{qu2022dgmil}, knowledge distillation frameworks~\cite{zhang2022dtfd,qu2022bi}, limited pathologist patch annotations~\cite{koga2025attention}, weak regional annotations~\cite{wang2022label}, and semi-supervised consistency regularization~\cite{jiang2023semi}. 
These hybrid strategies effectively generate additional supervision to refine instance predictions and leverage unlabeled data, boosting performance and data efficiency.
While some studies~\cite{liu2024pseudo,ouyang2024mergeup,aswolinskiy2025attention} explore mixup-like data augmentation between WSI pairs, supervision leveraging relationships across multiple slides is still unexplored.

\noindent\textbf{Batchsize in Computational Pathology.}
Batchsize is a crucial hyperparameter in deep learning. 
However, its exploration in CPath remains limited, primarily due to the computational demands of WSIs and the inherent data heterogeneity within each slide.
Consequently, mainstream slide-level MIL methods typically adopt a batchsize of 1~\cite{li2024generalizable,shi2024vila, li2024dynamic,song2024morphological,zhang2024attention,li2024rethinking,fourkioti2023camil,tang2025multipleinstancelearningframework,tang2025revisitingendtoendlearningslidelevel}.
For instance, RRTMIL~\cite{tang2024feature} utilizes slide-wise regional and cross-region self-attention to capture patch ordinality and heterogeneity within each slide, necessitating a batchsize of 1 to maintain intra-slide relationships.
Despite its prevalence, this practice often results in training instability and slow convergence, prompting methods such as gradient accumulation over multiple slides~\cite{koga2025attention,zhang2025icfnet} or instance-level sampling strategies that select fixed-size subsets of patches per slide to mitigate computational overhead and improve learning stability~\cite{campanella2019clinical,liu2024attention}. 
Current slide-level methods treat batched inputs as an efficiency trade-off rather than a genuinely effective training strategy.
In this paper, we investigate the benefits of pack-based batched training for slide‑level prediction and explore practical guidelines in CPath. \textcolor{cvprblue}{\textbf{Supplementary}} gives more discussion about this topic and pack-based training.

\section{Method}

\subsection{MIL-based Computational Pathology}
\label{sec:mil_pre}

\begin{figure}
  \vspace{-0.5cm}
  \centering
  \includegraphics[width=1.\linewidth]{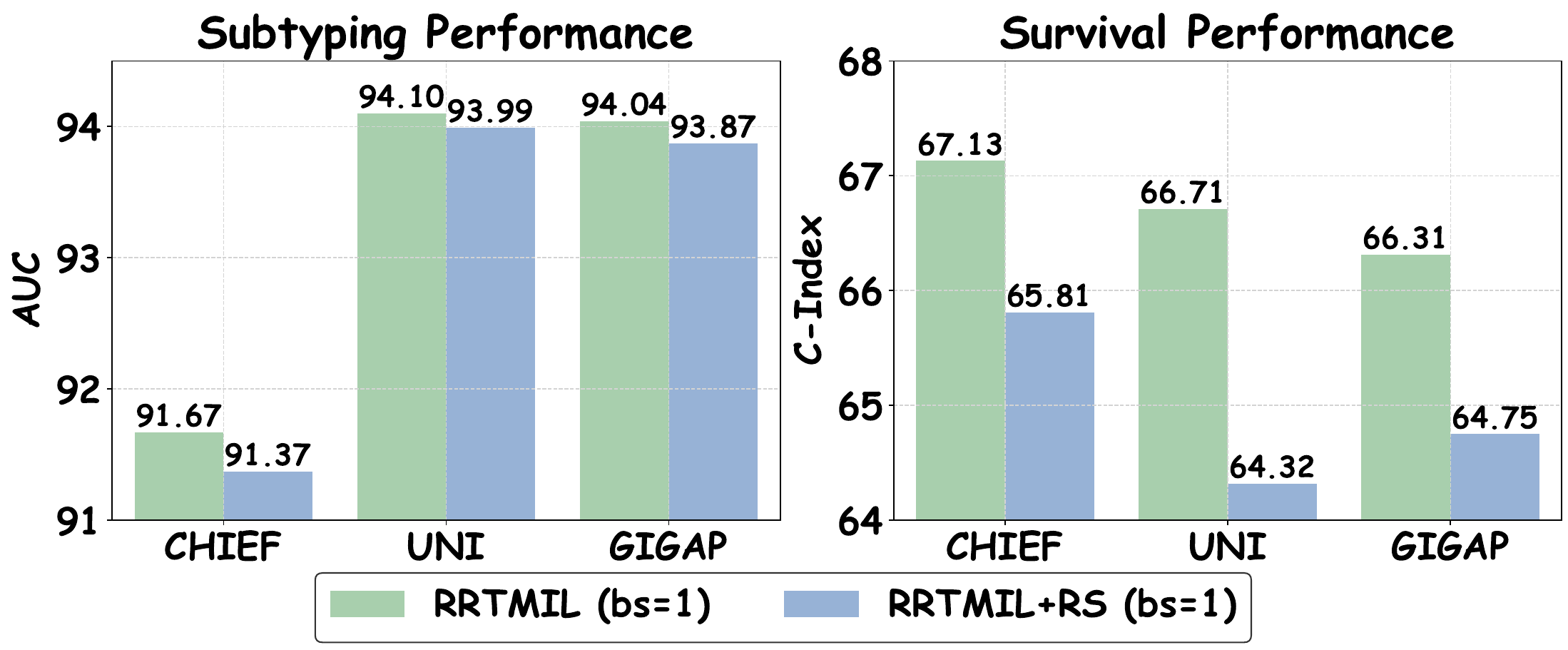}
  \caption{Impact of data heterogeneity on CPath. RS represents random sampling instances in all bags to a fixed length while maintaining original label, thus losing data heterogeneity.}
  \label{fig:ps}
  \vspace{-0.3cm}
\end{figure}

Histopathological WSIs are often gigapixel resolution, making direct processing impractical. 
Current approaches typically use weakly supervised MIL, where a WSI is treated as a bag $\mathcal{X}=\{x_{1},\dots,x_{N}\}$ of instances. 
During training, only a slide‑level label $y$ is available. 
Each instance $x_i$ is encoded to an embedding $h_{i}=f(x_{i})$ using a offline feature extractor $f$. An aggregation function $\Gamma(\cdot)$ combines instance embeddings $\{h_i\}_{i=1}^N$ into a bag-level representation $z = \Gamma_{\theta}(\{h_i\}_{i=1}^{N})$. 
This representation $z$ is then used by a classifier $g_{\phi}$ to predict the slide label $p(y\,|\,\mathcal{X}) = g_{\phi}(z)$. 
The significant variation in instance count $N$ across WSIs contribute to data heterogeneity.
This heterogeneity along with its associated spatial and morphological context, is crucial for CPath, as shown in Fig.\ref{fig:ps}. Specifically, when all instances are randomly sampled to a fixed-length sequence, a latest method like RRTMIL~\cite{tang2024feature} exhibits consistent performance degradation on multiple benchmarks, particularly for complex tasks like survival analysis. 
To maintain data heterogeneity, current methods necessitates training with a batchsize of 1,
resulting in noisy gradient estimates and optimization instability.

\begin{figure*}
\centerline{\includegraphics[width=.9\textwidth]{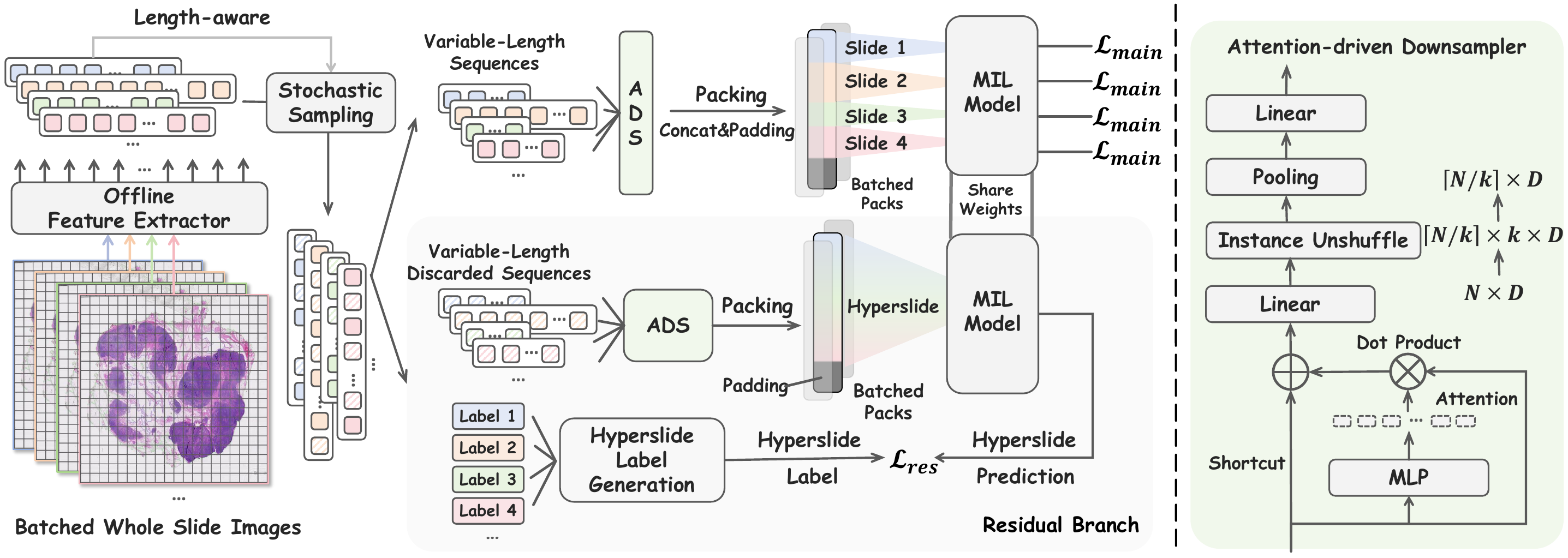}}
\caption{
\textbf{Left:} Overview of proposed pack-based MIL training framework. Instance features from each WSI are sampled into kept and discarded sequences. Both sequences are processed by ADS. Downsampled sequences from different bags are then concatenated into fixed-length packs. 
This packing mechanism aggregates different WSIs into fixed-length packs, thereby enabling batched training.
The dual-branch architecture, with shared weights, processes: 1) The \textit{Main Branch}, supervised by slide-level labels, 2) The \textit{Residual Branch}, supervised by Hyperslide labels. \textbf{Right:} Architecture of the Attention-driven Downsampler (ADS). 
Pseudo-codes are in \textcolor{cvprblue}{\textbf{Supplementary}}.
}
\label{fig:model}
\end{figure*}

\subsection{Pack-based MIL Training Framework}
CPath poses significant data challenges, including 
high data heterogeneity, redundancy and limited supervision, which hinder mainstream MIL method.
To overcome these issues, we propose a pack-based MIL training framework, named PackMIL.
As illustrated in Fig.\ref{fig:model}, the proposed method effectively employs a packing operation to maintain data heterogeneity and allow batched input, and it further provides multi-slide supervision, leading to more effective and stable training.
Consider a mini‑batch of $B$ bags, the $b$‑th bag provides a set of instance embeddings $\mathcal{H}_{b}=\{h_{bi}\}_{i=1}^{N_{b}}$ where $h_{bi}\!\in\!\mathbb{R}^{D}$.
However, the feature sequences extracted from gigapixel WSIs can be prohibitively long, making their direct inclusion into fixed-length packs impractical due to memory and computational limitations during training.
To address this, we employ stochastic instance-level sampling with ratio $r\in(0, 1)$.
The subsets $\mathcal{K}_b$ and $\mathcal{D}_b$ are formed as:
\begin{equation}
{\mathcal{K}}_{b}= \{\,h_{bi}\mid m_{bi}=1\,\},\quad
\mathcal{D}_{b}= \{\,h_{bi}\mid m_{bi}=0\,\},
\label{eq:dropout}
\end{equation}
where $m_{bi}\sim\operatorname{Bernoulli}(1-r)$.
We introduce the Residual Branch to mitigate feature loss during sampling and establish hyperslide supervision.
The kept sets $\mathcal{K}_b$ (main branch) and discarded sets $\mathcal{D}_b$ (residual branch) are processed to produce downsampled feature sets, $\tilde{\mathcal{K}}_b$ and $\tilde{\mathcal{D}}_b$, respectively.
To maintain data heterogeneity for batch training, features from sampled sets of each bag are sequentially arranged into fixed-length packs of size $L$. This packing operation $\operatorname{PACK}(\cdot)$ processes all bags in the mini-batch to form:
\begin{equation}
\begin{split}
\mathcal{P}^{\mathrm{main}} &= \operatorname{PACK}\Bigl(\{\tilde{\mathcal{K}}_b\}_{b=1}^{B},\,L\Bigr) \in\mathbb{R}^{B'\times L\times D}, \\
\mathcal{P}^{\mathrm{res}} &= \operatorname{PACK}\Bigl(\{\tilde{\mathcal{D}}_b\}_{b=1}^{B},\,L\Bigr) \in\mathbb{R}^{B''\times L\times D},
\end{split}
\label{eq:pack}
\end{equation}
where $\cup$ denotes concatenation along the instance axis
and $B'$ and $B''$ are the number of packs generated for the main and residual branches, respectively, calculated from the number of instances and $L$.
To ensure adequate representation of each slide, we enforce a minimum number of patches per slide in each pack. 
Zero-padding is applied in each pack when it contains fewer than $L$ patches.

We also incorporate an attention-driven downsampler (ADS) module for handling input redundancy between the two branches. 
This module fuses features from instances in $\mathcal{K}_b$ and $\mathcal{D}_b$ to generate more compact and informative feature representations.
Each set $\mathcal{S} \in \{\mathcal{K}_b, \mathcal{D}_b\}$ is downsampled by a factor $k\in\mathbb{N}$ using an ADS module, yielding $\tilde{\mathcal{S}}=\operatorname{ADS}(\mathcal{S}; k)$ with size $|\tilde{\mathcal{S}}|=\lceil|\mathcal{S}|/k\rceil$.

After packing, the main and residual branches are processed independently while sharing the same MIL model weight. 
In the main branch, each pack contains features originating from $B$ distinct bags. 
We use a mask $\mathbf{M}_p$ to identify the source bag for each instance feature within $p$-th pack. 
The embedding for the $b$-th bag is then computed by aggregating its instance features from the pack: 
$z^{\mathrm{main}}_{b}= \Gamma_\theta\!\Bigl( \mathbf{M}_{pb}, \mathcal{P}^{\mathrm{main}}_{p}\Bigr)$. 
We compute the main loss $\mathcal{L}_{\text{main}}$ 
over $B$ slides, based on slide-level predictions $\hat{y}_{b}=g_{\phi}(z^{\mathrm{main}}_{b})$. 
For the residual branch, each pack $p$ acts as a hyperslide, introducing high-level supervision. We compute its embedding $z^{\mathrm{res}}_{p}= \Gamma_\theta\!\Bigl( \mathcal{P}^{\mathrm{res}}_{p}\Bigr)$, obtain pack-level predictions $\hat{y}^{\mathrm{hyper}}_{p}=g_{\phi}(z^{\mathrm{res}}_{p})$, and compute the residual loss $\mathcal{L}_{\text{res}}$ 
over $B''$ packs.
The overall objective function is a weighted sum of the two losses: 
$\mathcal{L}= \mathcal{L}_{\text{main}} + \lambda\,\mathcal{L}_{\text{res}}$.

\noindent\textbf{Packing.}
The packing operation processes concatenated downsampled embeddings, $\tilde{\mathcal{K}}_b$ and $\tilde{\mathcal{D}}_b$.
The operation sequentially fills packs of fixed length $L$. Let $\mathcal{P}_p \in \mathbb{R}^{L \times D}$ be the $p$-th pack, for $p=1, \dots, B'$. Embeddings 
are placed sequentially into $\mathcal{P}_1, \mathcal{P}_2, \dots, \mathcal{P}_{B'}$. 
When a pack $\mathcal{P}_p$ cannot accommodate the next feature without exceeding $L$, or when all feature have been placed, any remaining positions in $\mathcal{P}_p$ are filled with vectors $\mathbf{0} \in \mathbb{R}^D$. 
To enforce the fixed pack length $L$, any downsampled features still exceeding this length are truncated to $L$.
Packing pseudocode is provided in \textcolor{cvprblue}{\textbf{Supplementary}}.

\noindent\textbf{Isolated Mask.}
We employ auxiliary masks to process features within each pack $\mathcal{P}_p$.
These masks serve a dual purpose: 1) \textit{preserving bag integrity by preventing cross-bag interactions}, and 2) \textit{nullifying the impact of zero-padding}.
Based on the CPath pipeline above, masks are divided into \textit{aggregation-oriented} and \textit{classification-oriented}.
Aggregation masks constrain the feature aggregation stages, ensuring that computations within each pack only involve instance features from the same source bag.
Classification-oriented masks subsequently select non-padding features for the final prediction.
These tailored masks enable efficient and effective MIL within our pack-based framework.
Detailed definition is provided in \textcolor{cvprblue}{\textbf{Supplementary}}.

\begin{figure*}
\centerline{\includegraphics[width=1.\textwidth]{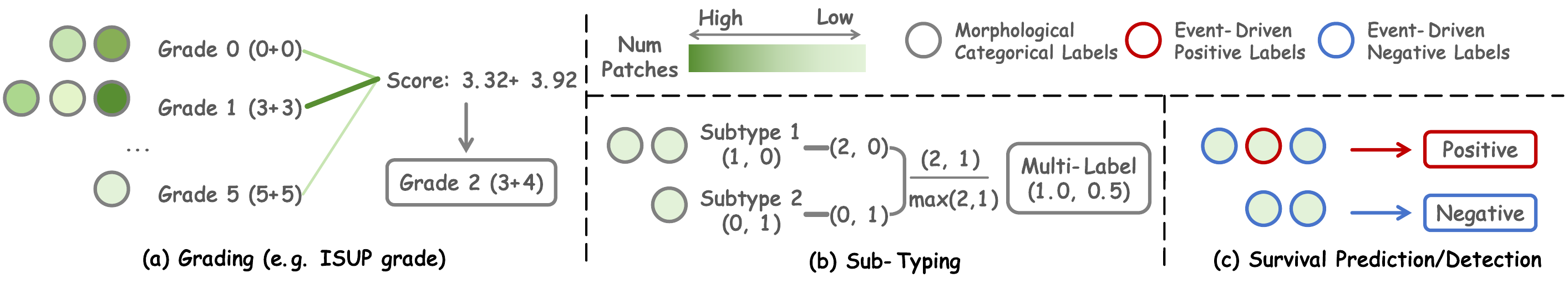}}
\caption{Illustration of Task-specific Hyperslide Labels. 
A \textit{hyperslide} is a conceptual pack formed by aggregating multiple WSIs to enable inter-slide supervision. These task-specific labels are constructed based on the clinical characteristics of different clinical tasks.}
\label{fig:label}
\end{figure*}

\noindent\textbf{Attention-driven Downsampler (ADS)}. 
WSIs exhibit significant redundancy ~\cite{tang2023mhim,zhang2024attention}, which can also manifest between the feature sets derived from $\mathcal{K}_b$ and $\mathcal{D}_b$. To fuse features while mitigating redundancy in dual branches, we designed the ADS module. 
This module utilizes attention-driven downsampling to produce a compact and informative representation.
Given a set of $N$ embeddings $\{h_i\}_{i=1}^N$, ADS computes a per-instance normalized attention score via a shallow MLP and applies it as a residual weigh
$u_i=h_i + a_i\,h_i$, which preserves the original feature while enhancing it based on its importance.
Followed by a learnable linear layer, we compute $W^L\!\in\!\mathbb{R}^{D\times D}$, $v_i = u_i\,W^L$.
Then
, we perform instance unshuffle, which rearranges sequential instance features into group-based representation: 
\begin{equation}
[v_1,\dots,v_N] \in \mathbb{R}^{N \times D}
\;\xrightarrow{\;\text{unshuffle by }k\;}\;
U \in \mathbb{R}^{\lceil N/k\rceil \times k \times D}.
\label{eq:unshuffle}
\end{equation}
where $k$ is the downsampling factor. 
ADS performs pooling along the $k$ dimension of each pack, followed by a projection $W^P\!\in\!\mathbb{R}^{D\times D}$. ADS takes the input features $\{h_i\}$ and a factor $k$, resulting in the output sequence $\operatorname{ADS}(\{h_i\};k) = \{\tilde h_j\}_{j=1}^{\lceil N/k\rceil} \in \mathbb{R}^{\lceil N/k\rceil\times D}$. Each feature $\tilde h_j$ is computed as:
\begin{equation}
\tilde h_j = \bigl[\operatorname{Pool}(U_j, \text{dim}=1)\bigr]\,W^P,\quad
j=1,\dots,\lceil N/k\rceil.
\label{eq:ads}
\end{equation}
where $\operatorname{Pool}(\cdot)$ is either random or max pooling, and $W^P\!\in\!\mathbb{R}^{D\times D}$ is a projection matrix.
By weighting instances with attention scores ${a_i}$ and pooling across structured groups, ADS reduces the instance count while prioritizing regions of high clinical relevance
, as the attention is trained with task supervision.
ADS also maintains interpretability at inference time. Additional details, discussions on applicable boundaries, and pseudocode are in the \textcolor{cvprblue}{\textbf{Supplementary}}.

\noindent\textbf{Inference Pipeline.} 
For inference, we adopt a deterministic pipeline to ensure reproducibility. The residual branch and stochastic sampling are bypassed. Each slide is processed individually with batchsize of 1, 
feeding its complete sequence of instance embeddings into the main branch.

\subsection{Training Recipe}
\textbf{Task-specific Hyperslide Labels and Loss.}
Heterogeneous pathological data and diverse downstream clinical tasks necessitate task-specific hyperslide labeling strategies that preserve task-relevant clinical characteristics and enable efficient learning from multiple WSIs.
We divide downstream tasks into two classes: \textit{Morphological Categorical} and \textit{Event-Driven}. 
Based on pathological scenarios, we design three strategies to generate a hyperslide label $y^\mathrm{hyper}$ for higher-level supervision, illustrated in Fig.\ref{fig:label}.
Appropriate loss functions are selected for different tasks, as detailed in the \textcolor{cvprblue}{\textbf{Supplementary}}.

1) Grading.
Let each pack contain $S$ WSIs, each associated with a Gleason Score $g_s$ (determined by the area proportions of the primary and secondary Gleason patterns) and a patch count $n_s$.
We compute a pack-level statistic by
$\tilde{g}= \frac{\sum_{s=1}^{S} n_s\,g_s}{\sum_{s=1}^{S} n_s}$.
The weighted metric is then mapped to a discrete ISUP grade $y^\mathrm{hyper} \in \{1,\dots,G\}$, serving as the single categorical hyperslide label.
Weighting by $n_s$ ensures that the pack-level grading reflects the relative tissue coverage of each WSI, preventing smaller tissue regions from being overrepresented in the final assessment.

2) Sub-typing. 
Sub-typing is cast as a \textit{multi-label} problem in which a single pack may express several subtypes simultaneously.  
For each subtype $c\in\{1,\dots,C\}$ we first obtain a slide-level indicator $t_{s,c}\!\in\!\{0,1\}$ and aggregate groundtruth across the $S$ slides by  
$\hat{p}_c \;=\; \frac{\sum_{s=1}^{S} n_s\,t_{s,c}}{\sum_{s=1}^{S} n_s}$,
where $n_s$ denotes the patch count of slide $s$.  
To capture the relative prevalence of co-occurring subtypes within pack, we generate hyperslide soft label $\mathbf{y}^\mathrm{hyper}$ by normalizing aggregated values using maximum value across all subtypes: 
$y^\mathrm{hyper}_c = \frac{\hat{p}_c}{\max_{j \in \{1,\dots,C\}} \hat{p}_j}$,
resulting in $\mathbf{y}^\mathrm{hyper} = [y^\mathrm{hyper}_1, \dots, y^\mathrm{hyper}_C]^\top \in [0,1]^C$.

3) Survival Analysis and Detection.
These tasks are inherently event-driven: labels correspond to clinical events with an intrinsic priority (e.g., tumor overrides normal).  We preserve this hierarchy by scanning the slides in descending priority order and assigning the first event observed:
\begin{equation}
y^\mathrm{hyper} = \arg\max_{e\in\mathcal{E}} \Bigl[ \max_{s} \mathbf{1}\{\,e\text{ occurs in slide }s\,\} \Bigr],
\end{equation}
where $\mathcal{E}$ is the ordered event set.
This strategy ensures that each pack is uniquely associated with the most clinically significant event.
\begin{figure*}
\centerline{\includegraphics[width=.85\textwidth]{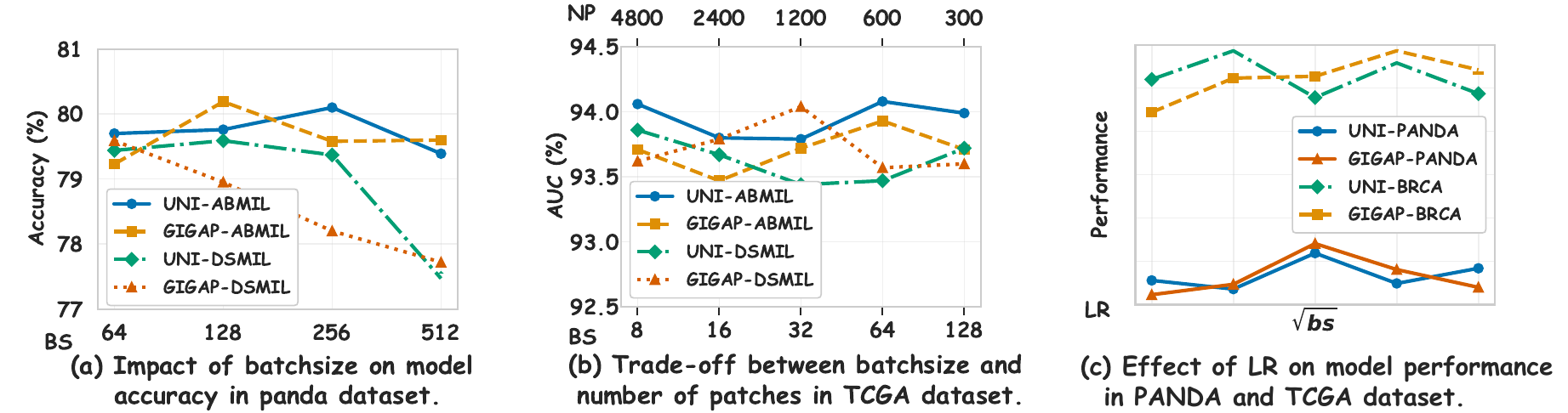}}
\caption{
\textbf{Practical Guidelines for Batched CPath Training.} 
Impact of training hyperparameters, highlighting dataset-scale-dependent strategies.
\textbf{(a)} On the large-scale PANDA dataset, accuracy is non-monotonic with $bs$, degrading at too large values. 
\textbf{(b)} On the smaller TCGA dataset, computational resource limitations necessitate an empirically tuned trade-off between $bs$ and number of instances. 
\textbf{(c)} The $\sqrt{bs}$ learning rate scaling rule is effective for PANDA (bottom) but fails on TCGA (top), showing standard rules are not universal in CPath.}
\label{fig:bs}
\end{figure*}

\noindent\textbf{Practical Guidelines for Batched Training.}
Selecting an appropriate batchsize ($bs$) across various CPath tasks remains an underexplored yet critical problem for effective model training. 
To establish an batched training protocol, we conduct a series of preliminary experiments, as shown in Fig.~\ref{fig:bs}.
The results from these experiments inform the practical guidelines discussed below, which are subsequently adopted in our main experimental setup. 
We find that the specifics of these guidelines vary significantly across benchmarks and are strongly correlated with dataset scale.

On conventional datasets (e.g., TCGA), which typically contain a limited number of WSIs
, resource constraints often force a trade-off between the $bs$ and the number of patches sampled per WSI (NP), as shown in Fig.~\ref{fig:bs}(b). The optimal choice of $bs$ usually requires empirical tuning. Additionally, learning rate is not scaled by regular rules and also requires empirical tuning, as depicted in Fig.~\ref{fig:bs}(c).

On large-scale datasets (e.g., PANDA~\cite{panda}), the $\sqrt{bs}$ learning rate scaling rule often proves effective. However, performance does not monotonically increase with $bs$; excessively large values can degrade performance, as illustrated in Fig.~\ref{fig:bs}(a). 
Moreover, we found that 1D Batch Normalization~\cite{ioffe2015batch} can effectively facilitate convergence with sufficient data scale. 
In contrast, normalization does not provide significant improvements on conventional CPath datasets, which we attribute to their limited data volume.

\section{Experiment} \label{sec:exp}

\subsection{Datasets and Evaluation Metrics}
For cancer diagnosis, we evaluate performance on grading and sub-typing tasks using the \textbf{PANDA~\cite{panda}} and \textbf{TCGA-BRCA} datasets. For cancer prognosis, we evaluate survival analysis performance using \textbf{TCGA-LUAD} and \textbf{TCGA-BRCA}. We report macro accuracy (Acc.) for cancer grading and the area under the ROC curve (AUC) for sub-typing. For survival analysis, we report the concordance index (C-index)~\cite{c-index_harrell}. To ensure statistical robustness, we perform 1000 bootstrap iterations and repeat 5 experiments for grading and subtyping, and 9 times for survival. We report the mean and 95\% confidence interval for all metrics. 
\textcolor{cvprblue}{\textbf{Supplementary}} also presents performance on the CAMELYON~\cite{c16} and fundus datasets.
\begin{table*}[t]
\scriptsize
\centering
\caption{Comparison of Grading (Acc.) on PANDA and Sub-typing (AUC) on BRCA. \textcolor{cvprblue}{\textbf{Supplementary}} gives 95\% CI of RS and PackMIL.}
\begin{tabular}{p{3cm}cccccc}
\toprule
\multirow{2}{*}{Method} 
& \multicolumn{3}{c}{Grading (Acc.$\uparrow$)} 
& \multicolumn{3}{c}{Sub-typing (AUC$\uparrow$)} \\ 
\cmidrule(lr){2-4} \cmidrule(lr){5-7}
& CHIEF (27M) & UNI (307M) & GIGAP (1134M)  
& CHIEF & UNI & GIGAP \\
\midrule
ABMIL~\cite{ilse2018attention} 
& 65.48$_{\pm0.70}$ & 73.21$_{\pm0.61}$ & 72.91$_{\pm0.62}$ 
& 89.58$_{\pm5.15}$ & 93.58$_{\pm3.92}$ & 94.13$_{\pm3.88}$ \\
DSMIL~\cite{li2021dual} 
& 71.95$_{\pm0.72}$ & 71.41$_{\pm0.68}$ & 70.84$_{\pm0.73}$ 
& 90.68$_{\pm4.91}$ & 93.89$_{\pm3.79}$ & 93.70$_{\pm4.14}$ \\
CLAM~\cite{clam}    
& 66.91$_{\pm0.73}$ & 74.76$_{\pm0.61}$ & 74.99$_{\pm0.60}$ 
& 89.61$_{\pm5.05}$ & 93.90$_{\pm3.86}$ & 93.73$_{\pm3.75}$ \\
TransMIL~\cite{shao2021transmil} 
& 68.53$_{\pm0.79}$ & 72.59$_{\pm0.75}$ & 71.91$_{\pm0.74}$
& 91.91$_{\pm4.34}$ & 94.09$_{\pm3.79}$ & 93.57$_{\pm3.84}$ \\
DTFD~\cite{zhang2022dtfd} 
& 63.57$_{\pm0.71}$ & 72.19$_{\pm0.66}$ & 71.93$_{\pm0.65}$ 
& 91.07$_{\pm4.59}$ & 93.86$_{\pm3.93}$ & 93.88$_{\pm3.81}$ \\
WiKG~\cite{li2024dynamic} 
& 72.37$_{\pm0.72}$ & 76.68$_{\pm0.68}$ & 74.91$_{\pm0.68}$ 
& 91.93$_{\pm4.65}$ & 93.81$_{\pm4.25}$ & 94.06$_{\pm3.92}$ \\
RRTMIL~\cite{tang2024feature} 
& 69.99$_{\pm0.69}$ & 72.18$_{\pm0.67}$ & 72.34$_{\pm0.63}$ 
& 91.67$_{\pm4.77}$ & 94.10$_{\pm3.65}$ & 94.04$_{\pm3.86}$ \\
2DMamba~\cite{zhang20242dmamba}
& 71.59$_{\pm0.73}$ & 74.97$_{\pm0.68}$ & 75.36$_{\pm0.66}$ 
& 90.96$_{\pm3.85}$ & 93.59$_{\pm3.14}$ & 93.33$_{\pm3.25}$ \\
\midrule
ABMIL\textcolor{olive}{+RS}
& 74.72 {\scriptsize\textcolor{kaiming-green}{+9.2}}
& 77.91 {\scriptsize\textcolor{kaiming-green}{+4.7}}
& 78.97 {\scriptsize\textcolor{kaiming-green}{+6.1}}
& 88.72 {\scriptsize\textcolor{red}{–0.9}}
& 93.89 {\scriptsize\textcolor{kaiming-green}{+0.3}}
& 93.78 {\scriptsize\textcolor{red}{-0.4}}\\
\rowcolor{gray}ABMIL\textcolor{olive}{+PackMIL} (Ours)
& \textbf{76.46 {\scriptsize\textcolor{kaiming-green}{+11.0}}}
& \textbf{80.19 {\scriptsize\textcolor{kaiming-green}{+7.0}}}
& \textbf{80.41 {\scriptsize\textcolor{kaiming-green}{+7.5}}}
& 92.38 {\scriptsize\textcolor{kaiming-green}{+2.8}}
& \textbf{94.86 {\scriptsize\textcolor{kaiming-green}{+1.3}}}
& \textbf{94.86 {\scriptsize\textcolor{kaiming-green}{+0.7}}}\\
\midrule
DSMIL\textcolor{olive}{+RS}    
& 75.00 {\scriptsize\textcolor{kaiming-green}{+3.1}} 
& 78.59 {\scriptsize\textcolor{kaiming-green}{+7.2}}
& 78.60 {\scriptsize\textcolor{kaiming-green}{+7.8}}
& 91.62 {\scriptsize\textcolor{kaiming-green}{+0.9}} 
& 93.29 {\scriptsize\textcolor{red}{–0.6}} 
& 94.04 {\scriptsize\textcolor{kaiming-green}{+0.3}} \\
\rowcolor{gray}DSMIL\textcolor{olive}{+PackMIL} (Ours)    
& 75.84 {\scriptsize\textcolor{kaiming-green}{+3.9}} 
& 79.68 {\scriptsize\textcolor{kaiming-green}{+8.3}} 
& 79.10 {\scriptsize\textcolor{kaiming-green}{+8.3}} 
& \textbf{93.01 {\scriptsize\textcolor{kaiming-green}{+2.3}}} 
& 94.62 {\scriptsize\textcolor{kaiming-green}{+0.7}} 
& 94.65 {\scriptsize\textcolor{kaiming-green}{+1.0}} \\
\midrule
TransMIL\textcolor{olive}{+RS}
& 73.68 {\scriptsize\textcolor{kaiming-green}{+5.2}}
& 76.94 {\scriptsize\textcolor{kaiming-green}{+4.4}}
& 76.15 {\scriptsize\textcolor{kaiming-green}{+4.2}}  
& 90.75 {\scriptsize\textcolor{red}{–1.2}} 
& 94.07 {\scriptsize\textcolor{red}{–0.0}} 
& 93.95 {\scriptsize\textcolor{kaiming-green}{+0.4}} \\
\rowcolor{gray}TransMIL\textcolor{olive}{+PackMIL} (Ours)
& 74.75 {\scriptsize\textcolor{kaiming-green}{+6.2}} 
& 78.87 {\scriptsize\textcolor{kaiming-green}{+6.3}} 
& 78.88 {\scriptsize\textcolor{kaiming-green}{+7.0}} 
& 92.31 {\scriptsize\textcolor{kaiming-green}{+0.4}} 
& 94.37 {\scriptsize\textcolor{kaiming-green}{+0.3}} 
& 94.12 {\scriptsize\textcolor{kaiming-green}{+0.6}} \\
\midrule
RRTMIL\textcolor{olive}{+RS}  
& 70.32 {\scriptsize\textcolor{kaiming-green}{+0.3}} 
& 75.04 {\scriptsize\textcolor{kaiming-green}{+2.9}} 
& 75.13 {\scriptsize\textcolor{kaiming-green}{+2.8}} 
& 91.55 {\scriptsize\textcolor{red}{–0.1}} 
& 94.02 {\scriptsize\textcolor{red}{-0.1}} 
& 93.70 {\scriptsize\textcolor{red}{-0.3}} \\
\rowcolor{gray}RRTMIL\textcolor{olive}{+PackMIL} (Ours)
& 74.63 {\scriptsize\textcolor{kaiming-green}{+4.6}} 
& 78.46 {\scriptsize\textcolor{kaiming-green}{+6.3}} 
& 78.43 {\scriptsize\textcolor{kaiming-green}{+6.1}} 
& 92.43 {\scriptsize\textcolor{kaiming-green}{+0.8}} 
& 94.54 {\scriptsize\textcolor{kaiming-green}{+0.4}} 
& 94.47 {\scriptsize\textcolor{kaiming-green}{+0.4}} \\
\bottomrule
\end{tabular}
\vspace{-0.2cm}
\label{tab:cls}
\end{table*}

\subsection{Main Results}
\noindent\textbf{Comparison Methods. }
We compare several established and MIL aggregators~\cite{ilse2018attention,clam,shao2021transmil,li2021dual,li2024dynamic,tang2024feature,zhang20242dmamba,zhang2022dtfd} using three SOTA pathology encoders: UNI~\cite{uni}, CHIEF~\cite{chief}, and GigaPath (GIGAP)~\cite{gigap}. 
\textcolor{cvprblue}{\textbf{Supplementary}} gives the additional comparisons with CONCHv1.5~\cite{lu2024avisionlanguage} and TITAN~\cite{ding2025multimodal}.
To comprehensively evaluate the proposed framework, we select four widely-used MIL models as baselines. Additionally, we compare PackMIL against a standard random sampling (\textcolor{olive}{RS}) strategy (i.e., sampling all inputs to a fixed length for batched training) to assess its effectiveness.

\begin{table*}[t]
\scriptsize
\centering
\caption{Comparison of Survival Analysis (C-index~\cite{c-index_harrell}) on BRCA and LUAD. OOM denotes Out-of-Memory in 24GB-RTX3090.}
\begin{tabular}{p{3cm}cccccc}
\toprule
\multirow{2}{*}{Method} 
& \multicolumn{3}{c}{Survival-BRCA (C-index$\uparrow$)} 
& \multicolumn{3}{c}{Survival-LUAD (C-index$\uparrow$)} \\ 
\cmidrule(lr){2-4} \cmidrule(lr){5-7}
& CHIEF (27M) & UNI (307M) & GIGAP (1134M)  
& CHIEF & UNI & GIGAP \\
\midrule
ABMIL~\cite{ilse2018attention} 
& 65.36$_{\pm9.00}$ & 65.87$_{\pm9.43}$ & 65.72$_{\pm8.86}$ 
& 61.99$_{\pm8.70}$ & 60.54$_{\pm8.83}$ & 59.88$_{\pm8.60}$ \\
DSMIL~\cite{li2021dual} 
& 65.81$_{\pm9.17}$ & 65.87$_{\pm9.98}$ & 64.94$_{\pm9.22}$ 
& 62.38$_{\pm8.50}$ & 61.44$_{\pm8.76}$ & 61.35$_{\pm8.20}$ \\
CLAM~\cite{clam}   
& 65.03$_{\pm9.38}$ & 65.45$_{\pm10.0}$ & 63.91$_{\pm9.70}$ 
& 61.78$_{\pm8.84}$ & 59.71$_{\pm8.48}$ & 60.70$_{\pm8.68}$ \\
TransMIL~\cite{shao2021transmil} 
& 65.75$_{\pm8.87}$ & 65.34$_{\pm9.44}$ & 65.35$_{\pm9.18}$ 
& 63.68$_{\pm8.66}$ & 62.63$_{\pm8.70}$ & 62.53$_{\pm8.53}$ \\
DTFD~\cite{zhang2022dtfd} 
& 67.22$_{\pm8.91}$ & 65.05$_{\pm10.3}$ & 65.54$_{\pm9.08}$ 
& 62.87$_{\pm8.56}$ & 60.83$_{\pm8.64}$ & 60.88$_{\pm8.67}$ \\
WIKG~\cite{li2024dynamic} 
& 65.55$_{\pm9.14}$ & 65.77$_{\pm9.21}$ & 65.79$_{\pm9.62}$ 
& OOM & OOM & OOM \\
RRTMIL~\cite{tang2024feature} 
& 67.13$_{\pm8.77}$ & 66.71$_{\pm9.95}$ & 66.31$_{\pm9.48}$ 
& 63.51$_{\pm8.76}$ & 61.32$_{\pm8.73}$ & 62.41$_{\pm8.41}$ \\
2DMamba~\cite{zhang20242dmamba} 
& 66.01$_{\pm7.06}$ & 65.73$_{\pm8.15}$ & 65.96$_{\pm7.69}$ 
& 63.23$_{\pm6.70}$ & 60.69$_{\pm6.97}$ & 62.19$_{\pm6.73}$ \\
\midrule
ABMIL\textcolor{olive}{+RS} 
& 64.02 {\scriptsize\textcolor{red}{–1.3}}
& 65.71 {\scriptsize\textcolor{red}{–0.2}}
& 63.88 {\scriptsize\textcolor{red}{–1.8}}
& 61.98 {\scriptsize\textcolor{red}{-0.0}}
& 60.34 {\scriptsize\textcolor{red}{–0.2}}
& 61.01 {\scriptsize\textcolor{kaiming-green}{+1.1}}\\
\rowcolor{gray}ABMIL\textcolor{olive}{+PackMIL} (Ours)
& 68.30 {\scriptsize\textcolor{kaiming-green}{+2.9}}
& 68.14 {\scriptsize\textcolor{kaiming-green}{+2.3}}
& 67.04 {\scriptsize\textcolor{kaiming-green}{+1.3}}
& 63.72 {\scriptsize\textcolor{kaiming-green}{+1.7}}
& 62.60 {\scriptsize\textcolor{kaiming-green}{+2.1}}
& 61.58 {\scriptsize\textcolor{kaiming-green}{+1.7}}\\
\midrule
DSMIL\textcolor{olive}{+RS} 
& 65.00 {\scriptsize\textcolor{red}{–0.8}} 
& 64.72 {\scriptsize\textcolor{red}{–1.2}} 
& 65.69 {\scriptsize\textcolor{kaiming-green}{+0.8}}
& 63.28 {\scriptsize\textcolor{kaiming-green}{+0.9}}
& 61.53 {\scriptsize\textcolor{kaiming-green}{+0.1}} 
& 61.00 {\scriptsize\textcolor{red}{–0.4}} \\
\rowcolor{gray}DSMIL\textcolor{olive}{+PackMIL} (Ours)
& \textbf{69.76 {\scriptsize\textcolor{kaiming-green}{+4.0}}} 
& \textbf{70.00 {\scriptsize\textcolor{kaiming-green}{+4.1}}} 
& \textbf{68.03 {\scriptsize\textcolor{kaiming-green}{+3.1}}} 
& 64.10 {\scriptsize\textcolor{kaiming-green}{+1.7}} 
& 62.18 {\scriptsize\textcolor{kaiming-green}{+0.7}} 
& 62.44 {\scriptsize\textcolor{kaiming-green}{+1.1}} \\
\midrule
TransMIL\textcolor{olive}{+RS} 
& 66.39 {\scriptsize\textcolor{kaiming-green}{+0.6}} 
& 65.14 {\scriptsize\textcolor{red}{–0.2}} 
& 65.52 {\scriptsize\textcolor{kaiming-green}{+0.2}}
& 63.53 {\scriptsize\textcolor{red}{–0.2}} 
& 62.12 {\scriptsize\textcolor{red}{–0.5}} 
& 61.03 {\scriptsize\textcolor{red}{–1.5}} \\
\rowcolor{gray}TransMIL\textcolor{olive}{+PackMIL} (Ours)
& 68.08 {\scriptsize\textcolor{kaiming-green}{+2.3}} 
& 68.44 {\scriptsize\textcolor{kaiming-green}{+3.1}}
& 66.80 {\scriptsize\textcolor{kaiming-green}{+1.5}} 
& 64.01 {\scriptsize\textcolor{kaiming-green}{+0.3}} 
& \textbf{63.61 {\scriptsize\textcolor{kaiming-green}{+1.0}}} 
& \textbf{63.04 {\scriptsize\textcolor{kaiming-green}{+0.5}}} \\
\midrule
RRTMIL\textcolor{olive}{+RS} 
& 66.11 {\scriptsize\textcolor{red}{–1.0}}
& 64.68 {\scriptsize\textcolor{red}{–2.0}} 
& 65.19 {\scriptsize\textcolor{red}{–1.1}} 
& 62.52 {\scriptsize\textcolor{red}{–1.0}} 
& 61.44 {\scriptsize\textcolor{kaiming-green}{+0.1}}
& 61.35 {\scriptsize\textcolor{red}{–1.1}} \\
\rowcolor{gray}RRTMIL\textcolor{olive}{+PackMIL} (Ours)
& 68.15 {\scriptsize\textcolor{kaiming-green}{+1.0}} 
& 68.73 {\scriptsize\textcolor{kaiming-green}{+2.0}} 
& 67.62 {\scriptsize\textcolor{kaiming-green}{+1.3}} 
& \textbf{64.37 {\scriptsize\textcolor{kaiming-green}{+0.9}}} 
& 62.01 {\scriptsize\textcolor{kaiming-green}{+0.7}} 
& 62.79 {\scriptsize\textcolor{kaiming-green}{+0.4}} \\
\bottomrule
\end{tabular}
\vspace{-0.3cm}
\label{tab:surv}
\end{table*}

\noindent\textbf{Focusing on Data Challenges in the FM Era.} Although MIL architectures have advanced significantly, their performance improvements become marginal when using offline features extracted by foundation models (FMs). For example, with UNI features, the gap between the latest and classic MIL methods in TCGA-BRCA-subtyping is only 0.52\%. The quality of the FM primarily determines the final performance, and the latest or more complex MIL methods have reached a performance bottleneck~\cite{chen2024benchmarking,shaomultiple}. In this context, we observe that \textit{addressing the inherent data challenges in CPath is an effective way to achieve significant performance gains.}
Specifically, random sampling (RS) achieves substantial improvements on grading tasks (Tab.~\ref{tab:cls}), especially on large datasets such as PANDA ($\sim$10,000 slides).
We attribute this primarily to its effectiveness in reducing input redundancy and the advantages of batched training on large-scale data.  However, this RS strategy performs inconsistently on benchmarks like TCGA-BRCA-subtyping, which exhibits greater data heterogeneity (e.g., a sequence length variation of 60,000 compared to 1,000 in the PANDA dataset). It provides only marginal gains on some benchmarks while degrading performance on others in this task. As shown in Tab.~\ref{tab:surv}, this issue becomes more pronounced in survival analysis, where sequence length variation are even larger. 
This sensitivity to heterogeneity is further reflected in its performance on complex methods like RRTMIL, where RS shows a significant performance gap across all benchmarks.
These results indicate that RS compromises the WSI heterogeneity preserved during traditional training (batchsize = 1) and suffers from feature loss due to sampling.

\noindent\textbf{PackMIL More Comprehensively Alleviates Data Challenges.} Compared with RS, PackMIL more comprehensively alleviates the various data challenges in CPath, achieving more substantial and consistent performance gains (+1.28\% on TCGA-BRCA-subtyping). For example, its superior performance in survival analysis demonstrates its ability to handle data heterogeneity. Moreover, PackMIL yields significant further improvements even on grading tasks where RS already performed well. We attribute this gain to PackMIL's ability to mitigate challenges of insufficient supervision. Notably, by comprehensively addressing these data challenges, PackMIL enables models to achieve performance comparable to higher-quality features with lower-cost offline features. Furthermore, it achieves consistent improvements with better features, demonstrating its generalizability. In summary, results demonstrate the significant impact of data challenges on CPath performance and validate the effectiveness of PackMIL.

\subsection{Ablation Study}
Unless otherwise specified, our ablations use ABMIL as the baseline model and UNI as the offline feature extractor. For the survival analysis task, we utilize the larger BRCA dataset.
\textcolor{cvprblue}{\textbf{Supplementary}} gives further discussion for each subsection, qualitative analysis, inference efficiency and key hyperparameters (i.e., $k$, $\lambda$, and branch split ratio).

\begin{table*}[t]
    \centering
    \scriptsize
    \caption{
        \textbf{Top: }
        Ablation of PackMIL and computational cost on PANDA. TTime (RTX 3090 GPU-hours) denotes Train Time.
        Memory is the GPU memory which evaluated during training. 
        FPS stands for frames per second.
        ADS module is disabled by default on PANDA; we detail its computational cost in \textcolor{cvprblue}{\textbf{Supplementary}}.
        \textbf{Bottom: }Loss curves of main and hyperslide loss under different settings.
    }
    \label{tab:abl}
        \begin{tabular}{lccccccc}
            \toprule
            Method & Batched Training & TTime & Memory & FPS & Grad. & Sub. & Surv. \\ 
            \midrule
            ABMIL~\cite{ilse2018attention}  & {\ding{55}}  
            & 12h & 0.6G & 2056 & 73.21  & 93.58 & 65.87 \\
            \rowcolor{gray}
            PackMIL(AB.) (Ours) & $\checkmark$  
            & 4h & 2.8G & 1984 & \textbf{80.19}  & \textbf{94.86} & 68.14 \\
            TransMIL~\cite{shao2021transmil}   & {\ding{55}}
            & 55h & 1.1G & 142 & 72.59  & 94.09 & 65.34 \\  
            \rowcolor{gray}
            PackMIL(Trans.) (Ours) & $\checkmark$  
            & 6.5h & 4.4G & 731 & 78.87  & 94.37 & \textbf{68.44} \\
            \midrule
            ABMIL (Baseline)   & {\ding{55}}  
            & 12h & 0.6G & 2056 & 73.21  & 93.58 & 65.87 \\  
            + Random Sampling & $\checkmark$  
            & 2h & 1.3G & 2056 & 77.91  & 93.89 & 65.71 \\
            + Random Sampling + HyperSlide  & $\checkmark$  
            & 2.5h & 2.5G & 2056 & 79.93 & 94.02 &  67.15 \\  
            \midrule
            + Pack  & $\checkmark$  
            & 2.5h & 2.2G & 1984 & 79.02  & 94.06 & 66.15 \\
            + Pack + HyperSlide & $\checkmark$  
            & 4h & 2.8G & 1984 & 80.19  & 94.21 & 67.50 \\
            + Pack + HyperSlide + ADS& $\checkmark$  
            & - & - & - & -  & 94.86 & 68.14 \\  
            \bottomrule
        \end{tabular}
        \vspace*{0.5em}
        \includegraphics[width=0.85\textwidth]{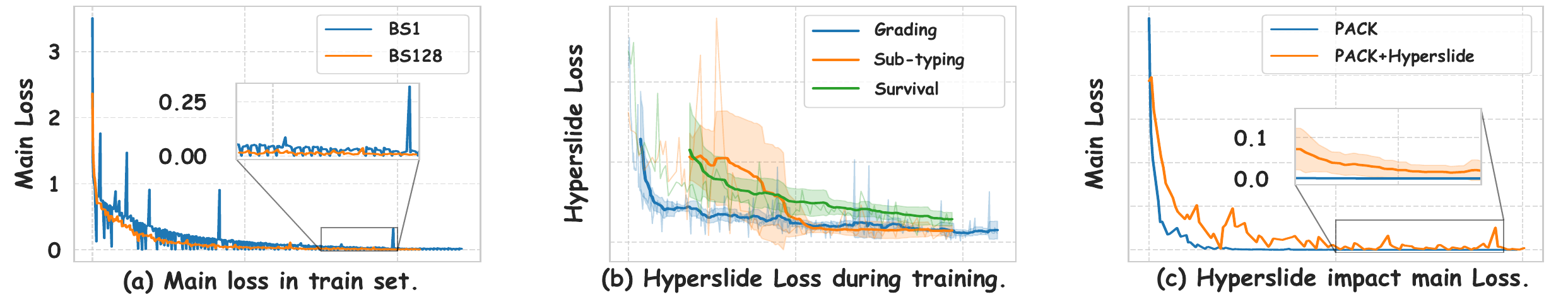}
        
        \vspace*{0.5em}
        \begin{tablenotes}[para, flushleft]
            \footnotesize
            \textbf{(a) Training loss convergence.} Batched training (\textcolor{orange}{orange}) exhibits faster and more stable convergence compared to the baseline with a batchsize of one (\textcolor{blue}{blue}).
            \textbf{(b) Auxiliary hyperslide loss.} The decreasing loss curve for the auxiliary hyperslide task demonstrates that the model effectively learns from the proposed multi-slide supervision.
            \textbf{(c) Impact of hyperslide on task loss.} The hyperslide (\textcolor{orange}{orange}) acts as a regularizer, mitigating the rapid overfitting (i.e., loss approaching zero in early epochs) observed when training with the task loss alone (\textcolor{blue}{blue}). 
            
        \end{tablenotes}
    \vspace*{-0.5em}
\end{table*}

\noindent\textbf{Batched Training and Packing Strategy. }
The results at the middle of Tab.~\ref{tab:abl} show that Random Sampling (RS) achieves significant improvements on the grading task, but its performance degrades on the other two tasks. This is attributed to its effective mitigation of input redundancy. However, this approach 
compromises data heterogeneity and leads to feature loss. 
Furthermore, batched training yields consistent performance improvements (row \textcolor{magenta}{6}). Importantly, it leads to significant gains in training stability and efficiency. Specifically, while maintaining faster and more stable convergence (Tab.~\ref{tab:abl}(a)), it also alleviates overfitting on the test set. 
With the pack-based batched training, PackMIL 
retains data heterogeneity. This allows PackMIL to benefit from batch processing while preserving the performance advantages of traditional training (batchsize=1).

\noindent\textbf{Hyperslide.}
We constructs a \textit{hyperslide} using discarded features in the residual branch and task-relevant hyperslide labels, aiming to offer multi-slide supervision while mitigating feature loss from sampling. Tab.~\ref{tab:abl}(b) confirms that the hyperslide learns effectively when guided by the proposed task-relevant labels. Furthermore, during joint training with the task loss, we find that optimizing the hyperslide also helps mitigate the rapid overfitting (i.e., the model rapid convergence of the training loss to near-zero on the training data within early epochs) of the model to the task loss on the FM features (Tab.~\ref{tab:abl}(c)). This issue hinders the model from benefiting from the task loss, consequently reducing training quality. The results in row \textcolor{magenta}{7} and row \textcolor{magenta}{9} of Tab.~\ref{tab:abl} demonstrate the performance improvements achieved by multi-slide supervision and minimizing feature loss.

\noindent\textbf{Attention-driven Downsampler.}
The Attention-driven Downsampler (ADS) is designed to mitigate potential input redundancy via feature fusion. Since the PANDA dataset contains fewer patches per WSI, we do not employ ADS on this task. However, for the sub-typing and survival tasks, which involve longer input sequences, the results indicate that ADS effectively reduces input redundancy. More important, through attention weighting and feature fusion, ADS preserves the discriminative information of the original features. 
\textcolor{cvprblue}{\textbf{Supplementary}} gives further discussion.

\noindent\textbf{Training Cost Analysis.}
On traditional small-scale datasets, CPath algorithms based on offline features rarely face significant efficiency challenges. However, this issue becomes increasingly prominent with the advent of large-scale datasets, such as PANDA.
Due to the non-batched training, 
we observe that training even a simple baseline like ABMIL on PANDA using offline features consumes 12 RTX3090 GPU-hours, while a more complex model like TransMIL requires 55 GPU-hours (top of Tab.~\ref{tab:abl}). These results highlight how non-batched training severely hinders the scalability of CPath algorithms to larger datasets. By enabling batched training, training time is significantly reduced (\textcolor{magenta}{-6$\sim$9x}), and accompanied by performance improvements. 
Notably, PackMIL achieves further significant performance gains with only a minor increase in computational overhead. It still maintaining a substantial efficiency advantage over non-batched training (\textcolor{magenta}{$\sim$12\%} training time).

\section{Discussion}
\noindent\textbf{Why Packing.}
Packing enables memory-efficient batched training while preserving the heterogeneity of WSI data. When dealing with significant sequence length variations (from 200 to 200k), padding is memory-inefficient, while sampling compromises data heterogeneity.

\noindent\textbf{Why Hyperslide.}
While constructing supervision labels within a single bag (intra-slide) is a mature approach in this field, Hyperslide packs instances from different bags to enable inter-slide supervision and knowledge supplementation. We believe this is a pioneering attempt that aligns with the clinical experience of pathologists.

\noindent\textbf{Packing Any MIL.}
PackMIL can be adapted to arbitrary MIL models. We have already implemented support for the most common and basic MIL architectures. Even if adapting certain complex architectures proves challenging, they can still be trained in a non-parallel fashion.
{
    \small
    \bibliographystyle{ieeenat_fullname}
    \bibliography{main}

@String(CVPR= {IEEE Conf. Comput. Vis. Pattern Recog.})

@String(ICASSP=	{ICASSP})

@String(CVPR  = {CVPR})

@inproceedings{tang2023mhim,
  title={Multiple Instance Learning Framework with Masked Hard Instance Mining for Whole Slide Image Classification},
  author={Tang, Wenhao and Huang, Sheng and Zhang, Xiaoxian and Zhou, Fengtao and Zhang, Yi and Liu, Bo},
  booktitle={Proceedings of the IEEE/CVF International Conference on Computer Vision},
  pages={4078--4087},
  year={2023}
}

@inproceedings{lin2023interventional,
  title={Interventional bag multi-instance learning on whole-slide pathological images},
  author={Lin, Tiancheng and Yu, Zhimiao and Hu, Hongyu and Xu, Yi and Chen, Chang-Wen},
  booktitle={Proceedings of the IEEE/CVF Conference on Computer Vision and Pattern Recognition},
  pages={19830--19839},
  year={2023}
}

@article{kingma2014adam,
  title={Adam: A method for stochastic optimization},
  author={Kingma, Diederik P and Ba, Jimmy},
  journal={arXiv preprint arXiv:1412.6980},
  year={2014}
}

@inproceedings{chen2022hipt,
  title={Scaling vision transformers to gigapixel images via hierarchical self-supervised learning},
  author={Chen, Richard J and Chen, Chengkuan and Li, Yicong and Chen, Tiffany Y and Trister, Andrew D and Krishnan, Rahul G and Mahmood, Faisal},
  booktitle={Proceedings of the IEEE/CVF Conference on Computer Vision and Pattern Recognition},
  pages={16144--16155},
  year={2022}
}

@article{maron1997mil_1,
  title={A framework for multiple-instance learning},
  author={Maron, Oded and Lozano-P{\'e}rez, Tom{\'a}s},
  journal={Advances in neural information processing systems},
  volume={10},
  year={1997}
}

@inproceedings{li2021dual,
  title={Dual-stream multiple instance learning network for whole slide image classification with self-supervised contrastive learning},
  author={Li, Bin and Li, Yin and Eliceiri, Kevin W},
  booktitle={CVPR},
  pages={14318--14328},
  year={2021}
}

@inproceedings{ilse2018attention,
  title={Attention-based deep multiple instance learning},
  author={Ilse, Maximilian and Tomczak, Jakub and Welling, Max},
  booktitle={ICML},
  pages={2127--2136},
  year={2018},
  organization={PMLR}
}

@article{clam,
  title={Data-efficient and weakly supervised computational pathology on whole-slide images},
  author={Lu, Ming Y and Williamson, Drew FK and Chen, Tiffany Y and Chen, Richard J and Barbieri, Matteo and Mahmood, Faisal},
  journal={Nature Biomedical Engineering},
  volume={5},
  number={6},
  pages={555--570},
  year={2021},
  publisher={Nature Publishing Group}
}

@article{c16,
  title={Diagnostic assessment of deep learning algorithms for detection of lymph node metastases in women with breast cancer},
  author={Bejnordi, Babak Ehteshami and Veta, Mitko and Van Diest, Paul Johannes and Van Ginneken, Bram and Karssemeijer, Nico and Litjens, Geert and Van Der Laak, Jeroen AWM and Hermsen, Meyke and Manson, Quirine F and Balkenhol, Maschenka and others},
  journal={Jama},
  volume={318},
  number={22},
  pages={2199--2210},
  year={2017},
  publisher={American Medical Association}
}

@article{shao2021transmil,
  title={Transmil: Transformer based correlated multiple instance learning for whole slide image classification},
  author={Shao, Zhuchen and Bian, Hao and Chen, Yang and Wang, Yifeng and Zhang, Jian and Ji, Xiangyang and others},
  journal={NeurIPS},
  volume={34},
  year={2021}
}

@inproceedings{zhang2022dtfd,
  title={DTFD-MIL: Double-Tier Feature Distillation Multiple Instance Learning for Histopathology Whole Slide Image Classification},
  author={Zhang, Hongrun and Meng, Yanda and Zhao, Yitian and Qiao, Yihong and Yang, Xiaoyun and Coupland, Sarah E and Zheng, Yalin},
  booktitle={Proceedings of the IEEE/CVF Conference on Computer Vision and Pattern Recognition},
  pages={18802--18812},
  year={2022}
}

@article{song2023artificial,
  title={Artificial intelligence for digital and computational pathology},
  author={Song, Andrew H and Jaume, Guillaume and Williamson, Drew FK and Lu, Ming Y and Vaidya, Anurag and Miller, Tiffany R and Mahmood, Faisal},
  journal={Nature Reviews Bioengineering},
  pages={1--20},
  year={2023},
  publisher={Nature Publishing Group UK London}
}

@article{cifci2023ai,
  title={AI in Computational Pathology of Cancer: Improving Diagnostic Workflows and Clinical Outcomes?},
  author={Cifci, Didem and Veldhuizen, Gregory P and Foersch, Sebastian and Kather, Jakob Nikolas},
  journal={Annual Review of Cancer Biology},
  volume={7},
  pages={57--71},
  year={2023},
  publisher={Annual Reviews}
}

@article{yao2020whole,
  title={Whole slide images based cancer survival prediction using attention guided deep multiple instance learning networks},
  author={Yao, Jiawen and Zhu, Xinliang and Jonnagaddala, Jitendra and Hawkins, Nicholas and Huang, Junzhou},
  journal={Medical Image Analysis},
  volume={65},
  pages={101789},
  year={2020},
  publisher={Elsevier}
}

@inproceedings{wen2023deep,
  title={Deep learning in digital pathology for personalized treatment plans of cancer patients},
  author={Wen, Zhuoyu and Wang, Shidan and Yang, Donghan M and Xie, Yang and Chen, Mingyi and Bishop, Justin and Xiao, Guanghua},
  booktitle={Seminars in Diagnostic Pathology},
  volume={40},
  number={2},
  pages={109--119},
  year={2023},
  organization={Elsevier}
}

@article{uni,
  title={Towards a general-purpose foundation model for computational pathology},
  author={Chen, Richard J and Ding, Tong and Lu, Ming Y and Williamson, Drew FK and Jaume, Guillaume and Song, Andrew H and Chen, Bowen and Zhang, Andrew and Shao, Daniel and Shaban, Muhammad and others},
  journal={Nature Medicine},
  volume={30},
  number={3},
  pages={850--862},
  year={2024},
  publisher={Nature Publishing Group US New York}
}

@article{gigap,
  title={A whole-slide foundation model for digital pathology from real-world data},
  author={Xu, Hanwen and Usuyama, Naoto and Bagga, Jaspreet and Zhang, Sheng and Rao, Rajesh and Naumann, Tristan and Wong, Cliff and Gero, Zelalem and Gonz{\'a}lez, Javier and Gu, Yu and others},
  journal={Nature},
  pages={1--8},
  year={2024},
  publisher={Nature Publishing Group UK London}
}

@article{chief,
  title={A pathology foundation model for cancer diagnosis and prognosis prediction},
  author={Wang, Xiyue and Zhao, Junhan and Marostica, Eliana and Yuan, Wei and Jin, Jietian and Zhang, Jiayu and Li, Ruijiang and Tang, Hongping and Wang, Kanran and Li, Yu and others},
  journal={Nature},
  volume={634},
  number={8035},
  pages={970--978},
  year={2024},
  publisher={Nature Publishing Group UK London}
}

@inproceedings{tang2024feature,
  title={Feature Re-Embedding: Towards Foundation Model-Level Performance in Computational Pathology},
  author={Tang, Wenhao and Zhou, Fengtao and Huang, Sheng and Zhu, Xiang and Zhang, Yi and Liu, Bo},
  booktitle={Proceedings of the IEEE/CVF Conference on Computer Vision and Pattern Recognition},
  pages={11343--11352},
  year={2024}
}

@inproceedings{li2024dynamic,
  title={Dynamic Graph Representation with Knowledge-aware Attention for Histopathology Whole Slide Image Analysis},
  author={Li, Jiawen and Chen, Yuxuan and Chu, Hongbo and Sun, Qiehe and Guan, Tian and Han, Anjia and He, Yonghong},
  booktitle={Proceedings of the IEEE/CVF Conference on Computer Vision and Pattern Recognition},
  pages={11323--11332},
  year={2024}
}

@inproceedings{fang2024sam,
  title={SAM-MIL: A Spatial Contextual Aware Multiple Instance Learning Approach for Whole Slide Image Classification},
  author={Fang, Heng and Huang, Sheng and Tang, Wenhao and Huangfu, Luwen and Liu, Bo},
  booktitle={Proceedings of the 32nd ACM International Conference on Multimedia},
  pages={6083--6092},
  year={2024}
}

@article{qu2022bi,
  title={Bi-directional weakly supervised knowledge distillation for whole slide image classification},
  author={Qu, Linhao and Wang, Manning and Song, Zhijian and others},
  journal={Advances in Neural Information Processing Systems},
  volume={35},
  pages={15368--15381},
  year={2022}
}

@article{panda,
  title={Artificial intelligence for diagnosis and Gleason grading of prostate cancer: the PANDA challenge},
  author={Bulten, Wouter and Kartasalo, Kimmo and Chen, Po-Hsuan Cameron and Str{\"o}m, Peter and Pinckaers, Hans and Nagpal, Kunal and Cai, Yuannan and Steiner, David F and Van Boven, Hester and Vink, Robert and others},
  journal={Nature medicine},
  volume={28},
  number={1},
  pages={154--163},
  year={2022},
  publisher={Nature Publishing Group US New York}
}

@inproceedings{song2024morphological,
  title={Morphological prototyping for unsupervised slide representation learning in computational pathology},
  author={Song, Andrew H and Chen, Richard J and Ding, Tong and Williamson, Drew FK and Jaume, Guillaume and Mahmood, Faisal},
  booktitle={Proceedings of the IEEE/CVF Conference on Computer Vision and Pattern Recognition},
  pages={11566--11578},
  year={2024}
}

@inproceedings{zhang2024attention,
  title={Attention-challenging multiple instance learning for whole slide image classification},
  author={Zhang, Yunlong and Li, Honglin and Sun, Yunxuan and Zheng, Sunyi and Zhu, Chenglu and Yang, Lin},
  booktitle={European Conference on Computer Vision},
  pages={125--143},
  year={2024},
  organization={Springer}
}

@article{li2024rethinking,
  title={Rethinking Transformer for Long Contextual Histopathology Whole Slide Image Analysis},
  author={Li, Honglin and Zhang, Yunlong and Chen, Pingyi and Shui, Zhongyi and Zhu, Chenglu and Yang, Lin},
  journal={arXiv preprint arXiv:2410.14195},
  year={2024}
}

@article{fourkioti2023camil,
  title={CAMIL: Context-aware multiple instance learning for cancer detection and subtyping in whole slide images},
  author={Fourkioti, Olga and De Vries, Matt and Jin, Chen and Alexander, Daniel C and Bakal, Chris},
  journal={arXiv preprint arXiv:2305.05314},
  year={2023}
}

@article{campanella2019clinical,
  title={Clinical-grade computational pathology using weakly supervised deep learning on whole slide images},
  author={Campanella, Gabriele and Hanna, Matthew G and Geneslaw, Luke and Miraflor, Allen and Werneck Krauss Silva, Vitor and Busam, Klaus J and Brogi, Edi and Reuter, Victor E and Klimstra, David S and Fuchs, Thomas J},
  journal={Nature medicine},
  volume={25},
  number={8},
  pages={1301--1309},
  year={2019},
  publisher={Nature Publishing Group US New York}
}

@article{pouransari2024dataset,
  title={Dataset decomposition: Faster llm training with variable sequence length curriculum},
  author={Pouransari, Hadi and Li, Chun-Liang and Chang, Jen-Hao and Anasosalu Vasu, Pavan Kumar and Koc, Cem and Shankar, Vaishaal and Tuzel, Oncel},
  journal={Advances in Neural Information Processing Systems},
  volume={37},
  pages={36121--36147},
  year={2024}
}

@article{krell2021efficient,
  title={Efficient sequence packing without cross-contamination: Accelerating large language models without impacting performance},
  author={Krell, Mario Michael and Kosec, Matej and Perez, Sergio P and Fitzgibbon, Andrew},
  journal={arXiv preprint arXiv:2107.02027},
  year={2021}
}

@article{brussee2025graph,
  title={Graph neural networks in histopathology: Emerging trends and future directions},
  author={Brussee, Siemen and Buzzanca, Giorgio and Schrader, Anne MR and Kers, Jesper},
  journal={Medical Image Analysis},
  pages={103444},
  year={2025},
  publisher={Elsevier}
}

@inproceedings{shao2023lnpl,
  title={Lnpl-mil: Learning from noisy pseudo labels for promoting multiple instance learning in whole slide image},
  author={Shao, Zhuchen and Wang, Yifeng and Chen, Yang and Bian, Hao and Liu, Shaohui and Wang, Haoqian and Zhang, Yongbing},
  booktitle={Proceedings of the IEEE/CVF international conference on computer vision},
  pages={21495--21505},
  year={2023}
}

@article{lin2025impact,
  title={Impact of stain variation and color normalization for prognostic predictions in pathology},
  author={Lin, Siyu and Zhou, Haowen and Watson, Mark and Govindan, Ramaswamy and Cote, Richard J and Yang, Changhuei},
  journal={Scientific Reports},
  volume={15},
  number={1},
  pages={2369},
  year={2025},
  publisher={Nature Publishing Group UK London}
}

@article{chen2022fast,
  title={Fast and scalable search of whole-slide images via self-supervised deep learning},
  author={Chen, Chengkuan and Lu, Ming Y and Williamson, Drew FK and Chen, Tiffany Y and Schaumberg, Andrew J and Mahmood, Faisal},
  journal={Nature Biomedical Engineering},
  volume={6},
  number={12},
  pages={1420--1434},
  year={2022},
  publisher={Nature Publishing Group UK London}
}

@misc{kaczmarzyk2024open,
  title={Open and reusable deep learning for pathology with WSInfer and QuPath. npj Precision Oncology 8, 1 (Jan. 2024)},
  author={Kaczmarzyk, Jakub R and O’Callaghan, Alan and Inglis, Fiona and Gat, Swarad and Kurc, Tahsin and Gupta, Rajarsi and Bremer, Erich and Bankhead, Peter and Saltz, Joel H},
  year={2024}
}

@inproceedings{wang2021hierarchical,
  title={Hierarchical graph pathomic network for progression free survival prediction},
  author={Wang, Zichen and Li, Jiayun and Pan, Zhufeng and Li, Wenyuan and Sisk, Anthony and Ye, Huihui and Speier, William and Arnold, Corey W},
  booktitle={Medical Image Computing and Computer Assisted Intervention--MICCAI 2021: 24th International Conference, Strasbourg, France, September 27--October 1, 2021, Proceedings, Part VIII 24},
  pages={227--237},
  year={2021},
  organization={Springer}
}

@article{eastwood2023mesograph,
  title={MesoGraph: Automatic profiling of mesothelioma subtypes from histological images},
  author={Eastwood, Mark and Sailem, Heba and Marc, Silviu Tudor and Gao, Xiaohong and Offman, Judith and Karteris, Emmanouil and Fernandez, Angeles Montero and Jonigk, Danny and Cookson, William and Moffatt, Miriam and others},
  journal={Cell Reports Medicine},
  volume={4},
  number={10},
  year={2023},
  publisher={Elsevier}
}

@inproceedings{qu2022dgmil,
  title={Dgmil: Distribution guided multiple instance learning for whole slide image classification},
  author={Qu, Linhao and Luo, Xiaoyuan and Liu, Shaolei and Wang, Manning and Song, Zhijian},
  booktitle={International conference on medical image computing and computer-assisted intervention},
  pages={24--34},
  year={2022},
  organization={Springer}
}

@article{koga2025attention,
  title={Attention induction based on pathologist annotations for improving whole slide pathology image classifier},
  author={Koga, Ryoichi and Yokota, Tatsuya and Arihiro, Koji and Hontani, Hidekata},
  journal={Journal of Pathology Informatics},
  volume={16},
  pages={100413},
  year={2025},
  publisher={Elsevier}
}

@article{wang2022label,
  title={Label cleaning multiple instance learning: Refining coarse annotations on single whole-slide images},
  author={Wang, Zhenzhen and Saoud, Carla and Wangsiricharoen, Sintawat and James, Aaron W and Popel, Aleksander S and Sulam, Jeremias},
  journal={IEEE transactions on medical imaging},
  volume={41},
  number={12},
  pages={3952--3968},
  year={2022},
  publisher={IEEE}
}

@article{jiang2023semi,
  title={A semi-supervised learning approach with consistency regularization for tumor histopathological images analysis},
  author={Jiang, Yanyun and Sui, Xiaodan and Ding, Yanhui and Xiao, Wei and Zheng, Yuanjie and Zhang, Yongxin},
  journal={Frontiers in Oncology},
  volume={12},
  pages={1044026},
  year={2023},
  publisher={Frontiers Media SA}
}

@article{zhang2025icfnet,
  title={ICFNet: Integrated Cross-modal Fusion Network for Survival Prediction},
  author={Zhang, Binyu and Meng, Zhu and Dong, Junhao and Su, Fei and Zhao, Zhicheng},
  journal={arXiv preprint arXiv:2501.02778},
  year={2025}
}

@article{liu2024attention,
  title={Attention is not what you need: Revisiting multi-instance learning for whole slide image classification},
  author={Liu, Xin and Zhang, Weijia and Zhang, Min-Ling},
  journal={arXiv preprint arXiv:2408.09449},
  year={2024}
}

@article{ciga2021overcoming,
  title={Overcoming the limitations of patch-based learning to detect cancer in whole slide images},
  author={Ciga, Ozan and Xu, Tony and Nofech-Mozes, Sharon and Noy, Shawna and Lu, Fang-I and Martel, Anne L},
  journal={Scientific Reports},
  volume={11},
  number={1},
  pages={8894},
  year={2021},
  publisher={Nature Publishing Group UK London}
}

@article{graham2019hover,
  title={Hover-net: Simultaneous segmentation and classification of nuclei in multi-tissue histology images},
  author={Graham, Simon and Vu, Quoc Dang and Raza, Shan E Ahmed and Azam, Ayesha and Tsang, Yee Wah and Kwak, Jin Tae and Rajpoot, Nasir},
  journal={Medical image analysis},
  volume={58},
  pages={101563},
  year={2019},
  publisher={Elsevier}
}

@article{liu2024pseudo,
  title={Pseudo-bag mixup augmentation for multiple instance learning-based whole slide image classification},
  author={Liu, Pei and Ji, Luping and Zhang, Xinyu and Ye, Feng},
  journal={IEEE Transactions on Medical Imaging},
  volume={43},
  number={5},
  pages={1841--1852},
  year={2024},
  publisher={IEEE}
}

@article{ouyang2024mergeup,
  title={MergeUp-augmented Semi-Weakly Supervised Learning for WSI Classification},
  author={Ouyang, Mingxi and Fu, Yuqiu and Yan, Renao and Shi, ShanShan and Ling, Xitong and Zhu, Lianghui and He, Yonghong and Guan, Tian},
  journal={arXiv preprint arXiv:2408.12825},
  year={2024}
}

@article{c-index_harrell,
  title={Multivariable prognostic models: issues in developing models, evaluating assumptions and adequacy, and measuring and reducing errors},
  author={Harrell Jr, Frank E and Lee, Kerry L and Mark, Daniel B},
  journal={Statistics in medicine},
  volume={15},
  number={4},
  pages={361--387},
  year={1996},
  publisher={Wiley Online Library}
}

@inproceedings{ioffe2015batch,
  title={Batch normalization: Accelerating deep network training by reducing internal covariate shift},
  author={Ioffe, Sergey and Szegedy, Christian},
  booktitle={International conference on machine learning},
  pages={448--456},
  year={2015},
  organization={pmlr}
}

@article{dehghani2023patch,
  title={Patch n’pack: Navit, a vision transformer for any aspect ratio and resolution},
  author={Dehghani, Mostafa and Mustafa, Basil and Djolonga, Josip and Heek, Jonathan and Minderer, Matthias and Caron, Mathilde and Steiner, Andreas and Puigcerver, Joan and Geirhos, Robert and Alabdulmohsin, Ibrahim M and others},
  journal={Advances in Neural Information Processing Systems},
  volume={36},
  pages={2252--2274},
  year={2023}
}

@article{wang2024qwen2,
  title={Qwen2-vl: Enhancing vision-language model's perception of the world at any resolution},
  author={Wang, Peng and Bai, Shuai and Tan, Sinan and Wang, Shijie and Fan, Zhihao and Bai, Jinze and Chen, Keqin and Liu, Xuejing and Wang, Jialin and Ge, Wenbin and others},
  journal={arXiv preprint arXiv:2409.12191},
  year={2024}
}

@article{zhang20242dmamba,
  title={2DMamba: Efficient State Space Model for Image Representation with Applications on Giga-Pixel Whole Slide Image Classification},
  author={Zhang, Jingwei and Nguyen, Anh Tien and Han, Xi and Trinh, Vincent Quoc-Huy and Qin, Hong and Samaras, Dimitris and Hosseini, Mahdi S},
  journal={arXiv preprint arXiv:2412.00678},
  year={2024}
}

@inproceedings{jaume2024transcriptomics,
  title={Transcriptomics-guided slide representation learning in computational pathology},
  author={Jaume, Guillaume and Oldenburg, Lukas and Vaidya, Anurag and Chen, Richard J and Williamson, Drew FK and Peeters, Thomas and Song, Andrew H and Mahmood, Faisal},
  booktitle={Proceedings of the IEEE/CVF Conference on Computer Vision and Pattern Recognition},
  pages={9632--9644},
  year={2024}
}

@inproceedings{li2024generalizable,
  title={Generalizable whole slide image classification with fine-grained visual-semantic interaction},
  author={Li, Hao and Chen, Ying and Chen, Yifei and Yu, Rongshan and Yang, Wenxian and Wang, Liansheng and Ding, Bowen and Han, Yuchen},
  booktitle={Proceedings of the IEEE/CVF Conference on Computer Vision and Pattern Recognition},
  pages={11398--11407},
  year={2024}
}

@inproceedings{shi2024vila,
  title={ViLa-MIL: Dual-scale Vision-Language Multiple Instance Learning for Whole Slide Image Classification},
  author={Shi, Jiangbo and Li, Chen and Gong, Tieliang and Zheng, Yefeng and Fu, Huazhu},
  booktitle={Proceedings of the IEEE/CVF Conference on Computer Vision and Pattern Recognition},
  pages={11248--11258},
  year={2024}
}

@inproceedings{zelasko2025emmett,
  title={Emmett: Efficient multimodal machine translation training},
  author={{\.Z}elasko, Piotr and Chen, Zhehuai and Wang, Mengru and Galvez, Daniel and Hrinchuk, Oleksii and Ding, Shuoyang and Hu, Ke and Balam, Jagadeesh and Lavrukhin, Vitaly and Ginsburg, Boris},
  booktitle={ICASSP 2025-2025 IEEE International Conference on Acoustics, Speech and Signal Processing (ICASSP)},
  pages={1--5},
  year={2025},
  organization={IEEE}
}

@article{kosec2021packing,
  title={Packing: Towards 2x nlp bert acceleration},
  author={Kosec, Matej and Fu, Sheng and Krell, Mario Michael},
  year={2021}
}

@article{bolya2022token,
  title={Token merging: Your vit but faster},
  author={Bolya, Daniel and Fu, Cheng-Yang and Dai, Xiaoliang and Zhang, Peizhao and Feichtenhofer, Christoph and Hoffman, Judy},
  journal={arXiv preprint arXiv:2210.09461},
  year={2022}
}

@inproceedings{ridnik2021asymmetric,
  title={Asymmetric loss for multi-label classification},
  author={Ridnik, Tal and Ben-Baruch, Emanuel and Zamir, Nadav and Noy, Asaf and Friedman, Itamar and Protter, Matan and Zelnik-Manor, Lihi},
  booktitle={Proceedings of the IEEE/CVF international conference on computer vision},
  pages={82--91},
  year={2021}
}

@inproceedings{lin2017focal,
  title={Focal loss for dense object detection},
  author={Lin, Tsung-Yi and Goyal, Priya and Girshick, Ross and He, Kaiming and Doll{\'a}r, Piotr},
  booktitle={Proceedings of the IEEE international conference on computer vision},
  pages={2980--2988},
  year={2017}
}

@article{tu2022dual,
  title={Dual-curriculum contrastive multi-instance learning for cancer prognosis analysis with whole slide images},
  author={Tu, Chao and Zhang, Yu and Ning, Zhenyuan},
  journal={Advances in neural information processing systems},
  volume={35},
  pages={29484--29497},
  year={2022}
}

@article{li2019diagnostic,
  title={Diagnostic assessment of deep learning algorithms for diabetic retinopathy screening},
  author={Li, Tao and Gao, Yingqi and Wang, Kai and Guo, Song and Liu, Hanruo and Kang, Hong},
  journal={Information Sciences},
  volume={501},
  pages={511--522},
  year={2019},
  publisher={Elsevier}
}

@article{litjens20181399,
  title={1399 H\&E-stained sentinel lymph node sections of breast cancer patients: the CAMELYON dataset},
  author={Litjens, Geert and Bandi, Peter and Ehteshami Bejnordi, Babak and Geessink, Oscar and Balkenhol, Maschenka and Bult, Peter and Halilovic, Altuna and Hermsen, Meyke and Van de Loo, Rob and Vogels, Rob and others},
  journal={GigaScience},
  volume={7},
  number={6},
  pages={giy065},
  year={2018},
  publisher={Oxford University Press}
}

@misc{ding2024titan,
      title={Multimodal Whole Slide Foundation Model for Pathology}, 
      author={Tong Ding and Sophia J. Wagner and Andrew H. Song and Richard J. Chen and Ming Y. Lu and Andrew Zhang and Anurag J. Vaidya and Guillaume Jaume and Muhammad Shaban and Ahrong Kim and Drew F. K. Williamson and Bowen Chen and Cristina Almagro-Perez and Paul Doucet and Sharifa Sahai and Chengkuan Chen and Daisuke Komura and Akihiro Kawabe and Shumpei Ishikawa and Georg Gerber and Tingying Peng and Long Phi Le and Faisal Mahmood},
      year={2024},
      eprint={2411.19666},
      archivePrefix={arXiv},
      primaryClass={eess.IV},
      url={https://arxiv.org/abs/2411.19666}, 
}

@article{lu2024avisionlanguage,
  title={A visual-language foundation model for computational pathology},
  author={Lu, Ming Y and Chen, Bowen and Williamson, Drew FK and Chen, Richard J and Liang, Ivy and Ding, Tong and Jaume, Guillaume and Odintsov, Igor and Le, Long Phi and Gerber, Georg and others},
  journal={Nature Medicine},
  pages={863–874},
  volume={30},
  year={2024},
  publisher={Nature Publishing Group}
}

@article{lu2024visual,
  title={A visual-language foundation model for computational pathology},
  author={Lu, Ming Y and Chen, Bowen and Williamson, Drew FK and Chen, Richard J and Liang, Ivy and Ding, Tong and Jaume, Guillaume and Odintsov, Igor and Le, Long Phi and Gerber, Georg and others},
  journal={Nature medicine},
  volume={30},
  number={3},
  pages={863--874},
  year={2024},
  publisher={Nature Publishing Group US New York}
}

@article{chen2024benchmarking,
  title={Benchmarking embedding aggregation methods in computational pathology: A clinical data perspective},
  author={Chen, Shengjia and Campanella, Gabriele and Elmas, Abdulkadir and Stock, Aryeh and Zeng, Jennifer and Polydorides, Alexandros D and Schoenfeld, Adam J and Huang, Kuan-lin and Houldsworth, Jane and Vanderbilt, Chad and others},
  journal={arXiv preprint arXiv:2407.07841},
  year={2024}
}

@inproceedings{shaomultiple,
  title={Do Multiple Instance Learning Models Transfer?},
  author={Shao, Daniel and Chen, Richard J and Song, Andrew H and Runevic, Joel and Lu, Ming Y and Ding, Tong and Mahmood, Faisal},
  booktitle={Forty-second International Conference on Machine Learning}
}

@inproceedings{zhang2025aem,
  title={AEM: attention entropy maximization for multiple instance learning based whole slide image classification},
  author={Zhang, Yunlong and Li, Honglin and Sun, Yuxuan and Shui, Zhongyi and Li, Jingxiong and Zhu, Chenglu and Yang, Lin},
  booktitle={International Conference on Medical Image Computing and Computer-Assisted Intervention},
  pages={45--55},
  year={2025},
  organization={Springer}
}

@inproceedings{dong2025fast,
  title={Fast and Accurate Gigapixel Pathological Image Classification with Hierarchical Distillation Multi-Instance Learning},
  author={Dong, Jiuyang and Jiang, Junjun and Jiang, Kui and Li, Jiahan and Zhang, Yongbing},
  booktitle={Proceedings of the Computer Vision and Pattern Recognition Conference},
  pages={30818--30828},
  year={2025}
}

@article{aswolinskiy2025attention,
  title={Attention-based whole-slide image compression achieves pathologist-level pre-screening of multi-organ routine histopathology biopsies},
  author={Aswolinskiy, Witali and van der Post, Rachel S and Campora, Michela and Baronchelli, Carla and Ardighieri, Laura and Vatrano, Simona and van der Laak, Jeroen and Munari, Enrico and Simons, Michiel and Nagtegaal, Iris and others},
  journal={Modern Pathology},
  pages={100827},
  year={2025},
  publisher={Elsevier}
}

@article{ding2025multimodal,
  title={A multimodal whole-slide foundation model for pathology},
  author={Ding, Tong and Wagner, Sophia J and Song, Andrew H and Chen, Richard J and Lu, Ming Y and Zhang, Andrew and Vaidya, Anurag J and Jaume, Guillaume and Shaban, Muhammad and Kim, Ahrong and others},
  journal={Nature Medicine},
  pages={1--13},
  year={2025},
  publisher={Nature Publishing Group US New York}
}

@misc{tang2025multipleinstancelearningframework,
      title={Multiple Instance Learning Framework with Masked Hard Instance Mining for Gigapixel Histopathology Image Analysis}, 
      author={Wenhao Tang and Sheng Huang and Heng Fang and Fengtao Zhou and Bo Liu and Qingshan Liu},
      year={2025},
      eprint={2509.11526},
      archivePrefix={arXiv},
      primaryClass={cs.CV},
      url={https://arxiv.org/abs/2509.11526}, 
}

@misc{tang2025revisitingendtoendlearningslidelevel,
      title={Revisiting End-to-End Learning with Slide-level Supervision in Computational Pathology}, 
      author={Wenhao Tang and Rong Qin and Heng Fang and Fengtao Zhou and Hao Chen and Xiang Li and Ming-Ming Cheng},
      year={2025},
      eprint={2506.02408},
      archivePrefix={arXiv},
      primaryClass={cs.CV},
      url={https://arxiv.org/abs/2506.02408}, 
}
}

\clearpage
\appendix
\addcontentsline{toc}{section}{Appendix}
\part{Appendix}
\begingroup
  \hypersetup{linkcolor=black}
  \parttoc
\endgroup

\newpage

\section{Datasets and Implementation Details}
\label{sec:app_did}
\subsection{Datasets}
We validate our method on various computational pathology tasks and challenging benchmarks in the ear of foundation models, including cancer grading (PANDA~\cite{panda}), subtyping (TCGA-BRCA), survival analysis (TCGA-LUAD, TCGA-BRCA).

\textbf{PANDA}~\cite{panda} is a large-scale, multi-center dataset dedicated to prostate cancer detection and grading. It comprises 10,202 digitized H\&E-stained whole-slide images, making it one of the most extensive public resources for prostate cancer histopathology. Each slide is annotated according to the Gleason grading system and subsequently assigned an International Society of Urological Pathology (ISUP) grade, enabling both cancer detection and severity assessment. 
Specifically, ISUP Grade 1 corresponds to Gleason 3+3, Grade 2 to 3+4, Grade 3 to 4+3, Grade 4 to Gleason score 8, and Grade 5 to Gleason score 9 or 10, while Grade 0 represents benign samples.
The dataset includes a diverse distribution of ISUP grades, with 2,724 slides classified as grade 0 (benign), 2,602 as grade 1, 1,321 as grade 2, 1,205 as grade 3, 1,187 as grade 4, and 1,163 as grade 5. Spanning multiple clinical centers, PANDA ensures a broad range of samples, mitigating center-specific biases.

The Breast Invasive Carcinoma (\textbf{TCGA-BRCA}) project is the sub-typing dataset we used. TCGA-BRCA includes two subtypes: Invasive Ductal Carcinoma (\textbf{IDC}) and Invasive Lobular Carcinoma (\textbf{ILC}). It contains 787 IDC slides and 198 ILC slides from 985 cases.

The primary goal of survival analysis is to estimate the survival probability or survival time of patients over a specific period. Therefore, we used the \textbf{TCGA-LUAD} and \textbf{TCGA-BRCA} projects to evaluate the model performance for survival analysis tasks. Unlike the diagnosis and sub-typing tasks, the survival analysis datasets are case-based rather than WSI-based. The WSIs of TCGA-BRCA are identical to those used in the sub-typing task but with different annotations. The TCGA-LUAD dataset includes 541 slides from 478 primarily Lung Adenocarcinoma cases.

\begin{table*}[tb]
\small
\centering
\caption{Quantitative comparison between existing packing methods, Neural Image Compression with Attention (NIC-A)~\cite{aswolinskiy2025attention}, and our proposed PackMIL framework across 12 benchmarks. Metrics reported are Accuracy for Grading, AUC for Sub-typing, and C-Index for Survival tasks.}
\label{tab:pack_comparison}
\resizebox{\textwidth}{!}{%
\begin{tabular}{lcccccccccccc}
\toprule
\multirow{2}{*}{\textbf{Method}} & \multicolumn{3}{c}{\textbf{PANDA} (Grading)} & \multicolumn{3}{c}{\textbf{BRCA} (Sub-typing)} & \multicolumn{3}{c}{\textbf{BRCA} (Survival)} & \multicolumn{3}{c}{\textbf{LUAD} (Survival)} \\
\cmidrule(lr){2-4} \cmidrule(lr){5-7} \cmidrule(lr){8-10} \cmidrule(lr){11-13}
 & CHIEF & UNI & GIGAP & CHIEF & UNI & GIGAP & CHIEF & UNI & GIGAP & CHIEF & UNI & GIGAP \\
\midrule
NIC-A~\cite{aswolinskiy2025attention} & 65.48 & 73.21 & 72.91 & 90.22 & 93.65 & 94.13 & 65.36 & 65.87 & 65.72 & 61.99 & 60.54 & 59.88 \\
\rowcolor{gray}PackMIL & \textbf{76.46} & \textbf{80.19} & \textbf{80.41} & \textbf{93.01} & \textbf{94.86} & \textbf{94.86} & \textbf{69.76} & \textbf{70.00} & \textbf{68.03} & \textbf{64.37} & \textbf{63.61} & \textbf{63.04} \\
\bottomrule
\end{tabular}%
}
\end{table*}

\noindent\textbf{Experiment Setting}. We randomly split the PANDA dataset into training, validation, and testing sets with a ratio of 7:1:2. Due to the limited data size, the remaining datasets are divided into training and testing sets with a ratio of 7:3. The grading and subtyping tasks use 5 different random seeds to ensure the stability of the results. And because the survival analysis task is more affected by data partitioning, we use 3-fold cross-validation with 3 different random seeds to conduct the experiments.

\subsection{Preprocess}
Following prior works~\cite{clam,shao2021transmil,zhang2022dtfd,tang2024feature}, we cropped each WSI into non-overlapping 256x256 patches at 20$\times$ magnification. As in CLAM~\cite{clam}, background regions, including holes, were discarded. The average number of patches is approximately 10,000 for TCGA and 500 for PANDA. To efficiently process the large number of patches, we adopted a traditional two-stage paradigm, employing pre-trained offline models for patch feature extraction. We utilized three state-of-the-art foundation models of varying sizes, pre-trained on WSIs: CHIEF~\cite{chief} (27M), UNI~\cite{uni} (307M), and GigaPath~\cite{gigap} (1134M). Their respective feature dimensions are 768, 1024, and 1536.

\subsection{Implementation Details}
\noindent\textbf{Training Details.}
For our experiments conducted with a batchsize of 1, which is a conventional approach for methods handling variable sequence lengths like the two-stage methods investigated~\cite{clam,shao2021transmil,tang2024feature}, we consistently employed the Adam optimizer~\cite{kingma2014adam}. An initial learning rate of $1\times 10^{-4}$ was used, and this rate was dynamically adjusted during training using the Cosine annealing strategy. 
To mitigate potential overfitting and ensure robust optimization, we incorporated an early stopping mechanism across all experiments, selecting the model checkpoint that achieved the best performance on the validation metric for final evaluation. For grading tasks, training ran for a maximum of 100 epochs with a patience of 20 (starting from epoch 75). For subtyping, the maximum was 75 epochs with a patience of 20 (starting from epoch 30). For survival analysis, the maximum was 100 epochs with a patience of 10 (starting from epoch 30). To ensure a fair comparison, all baseline methods were tuned following their official guidelines and recommended hyperparameter search spaces.
For experiments involving a batchsize greater than 1, the learning rate was dynamically adjusted to an empirically determined value to achieve optimal performance.
Notably, specific models encountered memory limitations on the 3090 GPU when applied to certain large datasets. For instance, training the WiKG aggregator~\cite{li2024dynamic} on the BRCA (Subtyping) dataset required sampling the number of patches down to 1024 per instance to fit into memory. 

\noindent\textbf{Training Resources.}
Except for the aforementioned cases, all experiments were performed on NVIDIA 3090 GPUs using unified hyperparameters where applicable.

\begin{table}[t]
    \centering
    \small
    \setlength{\tabcolsep}{4pt}
    \captionsetup{type=table}
    \caption{Details of different hyperparameters.}
    \begin{tabular}{lccc}
        \toprule
        Hyperparameter & Grad. & Sub. & Surv. \\
        \midrule
        \multicolumn{4}{l}{\textit{--- PackMIL ---}} \\
        Downsample Ratio $k$ & N/A & 4 & 3 \\
        Hyperslide-loss Weight $\lambda$ & 0.2 & 0.2 & 0.5 \\
        Branch split ratio & 40\% & 40\% & 50\% \\
        ADS-pooling & N/A & Random & Max \\
        \midrule
        \multicolumn{4}{l}{\textit{--- Random Sampling (RS) ---}} \\
        Sampling Number & 256 & 1300 & 1600 \\
        \bottomrule
    \end{tabular}
    \label{tab:hyper_value}
\end{table}

\noindent\textbf{Hyperparameters.}
Tab.~\ref{tab:hyper_value} gives the details of some important hyperparameters of PackMIL and Random Sampling (RS). Sec.~\ref{sec_hyper} gives more detailed discussion.

\section{Additional Quantitative Results}
\label{sec:app_abl}

\subsection{More Discussion about Packing Method}
While there have been preliminary explorations of packing strategies within computational pathology, these approaches remain nascent in scope and technical depth. Notably, methods such as Aswolinskiy et al.~\cite{aswolinskiy2025attention} rely on a coarse spatial packing technique, which merely stitches pixel-level tissue sections from the same case into a single macro-image to align with block-level labels. In stark contrast, our PackMIL framework introduces a sophisticated inter-slide-level packing mechanism that aggregates variable-length sequences from independent slides to resolve fundamental optimization bottlenecks. As evidenced by the quantitative comparisons in Table~\ref{tab:pack_comparison}, our approach significantly outperforms simple concatenation strategies, demonstrating the necessity of addressing data heterogeneity through robust feature-space optimization rather than naive spatial stitching.

\begin{table*}[h]
\small
\centering
\caption{Comparative experiments with advanced feature extractors and slide encoders.}
\label{tab:encoder_comparison}
\begin{tabular}{llcccc}
\toprule
Feature Extractor & Slide Encoder & Grading & Subtyping & Survival & TTime \\ \midrule
HIPT~\cite{chen2022hipt}                       & HIPT                   & 62.10 $\pm$ 1.06 & 86.93 $\pm$ 4.86   & 67.82 $\pm$ 9.61  & 32h            \\
GigaPath~\cite{gigap}                   & GigaPath               & 65.86 $\pm$ 0.77 & 93.72 $\pm$ 3.42   & 62.64 $\pm$ 9.33  & 50h            \\
GigaPath~\cite{gigap}                   & PackMIL                & \textbf{80.41 $\pm$ 0.53} & 94.86 $\pm$ 3.68   & 68.03 $\pm$ 9.10  & 3h             \\
CHIEF~\cite{chief}                      & CHIEF                  & 67.67 $\pm$ 0.84 & 91.43 $\pm$ 4.51   & 67.95 $\pm$ 8.46  & 15h            \\
CONCHv1.5~\cite{lu2024visual}                  & ABMIL                  & 66.90 $\pm$ 0.74 & 95.14 $\pm$ 2.92   & 68.65 $\pm$ 9.17  & 14h            \\
CONCHv1.5~\cite{lu2024visual}                   & TITAN                  & 63.72 $\pm$ 0.76 & 95.20 $\pm$ 2.92   & -                 & 75h            \\
CONCHv1.5~\cite{lu2024visual}                   & PackMIL                & 71.72 $\pm$ 0.65 & \textbf{95.43 $\pm$ 2.63}   & \textbf{70.74 $\pm$ 7.70}  & \textbf{2h}             \\ \bottomrule
\end{tabular}%
\end{table*}

\subsection{Additional Benchmarking Experiments}
\textbf{Additional Dataset.} To further substantiate the robustness and generalizability of our proposed PackMIL, we extended its evaluation to two additional, widely recognized benchmarks: a computational pathology task for cancer metastasis detection and a medical imaging task outside of pathology for diabetic retinopathy grading.
The cancer metastasis detection benchmark integrates the \textbf{CAMELYON-16} and \textbf{CAMELYON-17} datasets~\cite{litjens20181399}. These datasets consist of whole-slide images (WSIs) of hematoxylin and eosin (H\&E) stained lymph node sections from breast cancer patients. The primary task is to identify the presence of metastatic cancerous tissue within these lymph nodes, a critical step in cancer staging. For the non-pathology benchmark, we utilized a standard \textbf{Diabetic Retinopathy (DR) Grading} dataset~\cite{li2019diagnostic}. This dataset contains retinal fundus images and the objective is to classify them into different severity levels of diabetic retinopathy. This task serves to evaluate the model's applicability to broader medical image classification challenges beyond histopathology.
We compared PackMIL with several state-of-the-art Multiple Instance Learning (MIL) methods. As shown in Tab.~\ref{tab:benchmark_results}, PackMIL demonstrates superior performance on both benchmarks. Specifically, on the CAMELYON cancer metastasis detection task, PackMIL (AB.) achieved an accuracy of 98.42\%, outperforming all other methods. Similarly, in the Diabetic Retinopathy Grading task, PackMIL (AB.) obtained the highest score of 61.34\%. These results underscore the effectiveness and versatility of our approach across different medical imaging domains.

\begin{table}[t]
\small
\centering
\caption{Performance comparison on additional benchmark datasets.}
\label{tab:benchmark_results}
\begin{tabular}{lcc}
\toprule
\textbf{Method}      & \textbf{CAMELYON (UNI)} & \textbf{DR (R50)} \\ \midrule
ABMIL                & 96.58                   & 58.26                                       \\
DSMIL                & 96.44                   & 58.03                                       \\
TransMIL             & 96.63                   & 60.28                                       \\
WiKG                 & 97.08                   & 55.84                                       \\
PackMIL (AB.) & \textbf{98.42}          & \textbf{61.34}                              \\
PackMIL (DS.) & 97.81                   & 60.27                                       \\ \bottomrule
\end{tabular}
\end{table}

\noindent\textbf{Additional Encoders.}
To further assess the versatility and effectiveness of PackMIL, we conducted additional experiments by integrating it with more advanced feature extractors and comparing its performance against , powerful slide-level encoders. We utilized CONCHv1.5~\cite{lu2024avisionlanguage} as a feature extractor and compared our model with state-of-the-art slide encoders including HIPT~\cite{chen2022hipt}, GigaPath~\cite{gigap}, CHIEF~\cite{chief}, and TITAN~\cite{ding2024titan} on downstream tasks of tumor grading, subtyping, and survival prediction. We reported the training time (TTime) of all methods on PANDA.
The results, detailed in Tab.~\ref{tab:encoder_comparison}, demonstrate that PackMIL consistently enhances performance while maintaining remarkable efficiency. Notably, PackMIL significantly reduces the training time, requiring only 2-3 hours, which is a substantial improvement over the 15 to 75 hours required by other encoders. This highlights PackMIL's ability to effectively aggregate features from various powerful backbones, achieving superior predictive performance with significantly lower computational overhead.

\begin{table*}[t]
    \small
    \centering
    \caption{Ablation studies on various components of our method. Default settings are marked in \colorbox{gray}{gray}.}
    \begin{subtable}[t]{0.25\textwidth}
        \centering
        \begin{tabular}[t]{lcc} 
         & Sub. & Surv. \\ \toprule
        2       & 93.87     & 67.03    \\
        3       & 94.40     & \cellcolor{gray}68.14    \\
        4       & \cellcolor{gray}94.86     & 66.89    \\
        5       & 94.32     & 65.74    \\
        6       & 94.60     & 66.03    \\
        \end{tabular}
        \caption{\textbf{Downsample ratio $k$} of ADS.}
        \label{tab:abl1}
    \end{subtable}%
    \hfill%
    \begin{subtable}[t]{0.35\textwidth}
        \centering
        \begin{tabular}[t]{lccc}
        & Grad. & Sub. & Surv. \\ \toprule
        0.05    & 79.95     & 94.71     & 65.97    \\
        0.1     & 80.15     & 94.78     & 66.42    \\
        0.2     & \cellcolor{gray}80.19     & \cellcolor{gray}94.86     & 67.28    \\
        0.5     & 79.85     & 94.67     & \cellcolor{gray}68.14    \\
        1.0     & 79.94     & 94.56     & 69.31    \\
        \end{tabular}
        \caption{\textbf{hyperslide‐loss weight} $\lambda$.}
        \label{tab:abl2}
    \end{subtable}%
    \hfill%
    \begin{subtable}[t]{0.35\textwidth}
        \centering
        \begin{tabular}[t]{lccc}
        & Grad. & Sub. & Surv. \\ \toprule
        30\%       & 80.01     & 94.58     & 65.94    \\
        40\%       & \cellcolor{gray}80.19     & \cellcolor{gray}94.86     & 66.69    \\
        50\%       & 79.52     & 94.77     & \cellcolor{gray}68.14    \\
        60\%       & 79.04     & 94.86     & 68.23    \\
        \vspace{.0cm}
        \end{tabular}
        \caption{\textbf{Branch split ratio} of main branch.}
        \label{tab:abl3}
    \end{subtable}
    \label{tab:abl_sm}
\end{table*}
\subsection{Ablation of Hyperparameters}
\label{sec_hyper}
We conduct ablation studies on the hyperparameters related to our method in Tab.~\ref{tab:abl_sm} and provide the following analysis.

\noindent\textbf{Dowansmple Ratio of ADS. } The ADS downsampling ratio determines the final number of input instance, demonstrating different characteristics across tasks. For sub-typing tasks, a moderate or smaller number of instances (500–1500) is found to be feasible or even optimal. Consequently, larger downsampling ratios (which result in fewer instances) yielded good performance. However, for the more challenging survival analysis, the number of instances have a more significant impact, with a larger number of instances ($>2000$) often leading to better performance.

\noindent\textbf{Weight of Hyperslide Loss. } The influence of varying hyperslide-loss weight on overall optimization is examined. We observe that parameter choices within a certain range consistently provide substantial performance gains, indicating that this parameter is relatively stable. Furthermore, larger ratios are observed to yield better results specifically on the survival analysis task. This improved performance is likely due to the nature of the survival analysis task itself. Given its data volume and inherent difficulty compared to other tasks, the main loss function is more susceptible to overfitting on the training set, making the hyperslide optimization play a more critical role.

\noindent\textbf{Branch Split Ratio.} The Split ratio controls the proportion of instances in different branches. It is hypothesized that a relatively even distribution would be more conducive to the overall optimization of the dual-branch architecture. Experimental results support this hypothesis, as more uneven division ratios do not yield significant performance gains.

\subsection{More Discussion about Hyperslide}

\textbf{Effectiveness of the HyperSlide Supervision.}
To isolate the contribution of our multi-WSI supervision mechanism, we conducted an ablation study, the results of which are presented in Tab.~\ref{tab:hyperslide_ablation}. The study demonstrates that the introduction of HyperSlide labels provides a consistent and significant performance uplift across all three downstream tasks. This improvement holds true for both the baseline random sampling (RS) packing strategy and our proposed adaptive packing method. For instance, adding HyperSlide to our adaptive packing improved Grading AUC by 1.17, Subtyping AUC by 0.15, and Survival C-Index by 1.35. This underscores the efficacy of using a higher-level, clinically-informed supervision signal to guide the model's learning process on aggregated WSI data.

\begin{table}[t]
    \centering
    \small
    \setlength{\tabcolsep}{4pt}
    \captionsetup{type=table}
    \caption{Impact of HyperSlide supervision.}
    \begin{tabular}{lccc}
        \toprule
        Method / Task & Grading & Subtyping & Survival \\
        \midrule
        Random Sampling (RS) & 77.91 & 93.89 & 65.71 \\
        RS + HyperSlide & 79.93 & 94.02 & 67.15 \\
        \midrule
        Pack (Ours) & 79.02 & 94.06 & 66.15 \\
        \rowcolor{gray!20}
        Pack + HyperSlide (Ours) & \textbf{80.19} & \textbf{94.21} & \textbf{67.50} \\
        \bottomrule
    \end{tabular}
    \label{tab:hyperslide_ablation}
\end{table}

\begin{table}[hb]
    \small
    \centering
    \setlength{\tabcolsep}{4pt}
    \captionsetup{type=table}
    \caption{Comparison of task-specific HyperSlide labels and mixed soft-labels.}
    \begin{tabular}{lccc}
        \toprule
        Labeling Strategy / Task & Grading & Subtyping & Survival \\
        \midrule
        Mixed Soft Label & 75.50 & 91.31 & N/A \\
        \rowcolor{gray!20}
        Task-specific Label (Ours) & \textbf{80.19} & \textbf{94.21} & \textbf{67.50} \\
        \bottomrule
    \end{tabular}
    \label{tab:label_comparison}
\end{table}

\noindent\textbf{Superiority over Soft Labeling.}
The design of the HyperSlide label is critical to its success. We further compared our task-specific labeling strategies against a more naive `mixed soft label` baseline, which simply averages slide-level information without considering task-specific clinical nuances. As shown in Tab.~\ref{tab:label_comparison}, our carefully designed labels significantly outperform this simpler approach. The naive method not only yields substantially lower performance on categorical tasks but is also incompatible with event-driven tasks like survival analysis, where label priority is paramount. This result validates our core hypothesis: to effectively learn from multiple WSIs, the generated supervision signal must preserve the essential, task-relevant clinical characteristics of the slide ensemble.

\begin{table}[t]
    \centering
    \small
    \setlength{\tabcolsep}{4pt}
    \captionsetup{type=table}
    \caption{Extended comparison between batched training and gradient accumulation. Batched training not only accelerates training but is essential for the effectiveness of subsequent modules.}
    \begin{tabular}{lcccc}
        \toprule
        Strategy & Grad. & Sub. & Surv. & TTime\\
        \midrule
        \multicolumn{5}{l}{\textit{--- Without Batched Training ---}} \\
        Baseline & 73.21 & 93.58 & 65.87 & 12h \\
        Accumulation & 72.12 & 93.02 & 64.28 & 12h \\
        Accumulation + patch drop & 77.58 & 93.61 & 64.53 & 12h \\
        Accumulation + ADS & - & 94.25 & 66.98 & - \\
        \midrule
        \multicolumn{5}{l}{\textit{--- With Batched Training ---}} \\
        Batched (RS) + patch drop & 77.91 & 93.89 & 65.71 & 2h \\
        Batched (RS) + ADS & - & 94.27 & 67.04 & - \\
        Batched (RS) + HyperSlide & 79.93 & 94.02 & 67.15 & 2.5h \\
        \rowcolor{gray} 
        Ours & \textbf{80.19} & \textbf{94.86} & \textbf{68.14} & 4h \\
        \bottomrule
    \end{tabular}
    \label{tab:ga_extended}
\end{table}

\subsection{More Discussion about Batched Training}

\textbf{Batched training is superior to gradient accumulation.}
Gradient accumulation is often considered an alternative to batched learning, aiming to simulate larger batch sizes while preserving data heterogeneity. However, our experiments demonstrate its inferiority in CPath tasks that utilize features from foundation models. 
As detailed in Tab.~\ref{tab:ga_extended}, the standard gradient accumulation strategy not only failed to improve training efficiency (maintaining a 12-hour training time) but also consistently underperformed the simple batched training baseline. More importantly, we found that batched training is a crucial prerequisite for unlocking the performance gains from subsequent optimization modules. When combined with techniques like patch dropout or ADS, the batched training approach consistently outperforms its gradient accumulation counterpart. This suggests that the batch-level feature interaction is vital for these modules to function effectively.
Furthermore, batched training provides a significant training speedup (up to $8\times$), reducing training time from 12 hours to under 3 hours in most configurations. This efficiency is critical for iterative research and large-scale experimentation.
This evidence underscores that the dual benefits of batched training, namely its efficiency and its role as a foundation for advanced modeling, are difficult to replace. In contrast, our proposed packing strategy builds upon the efficiency of batched training. It further enhances data heterogeneity and enables the crucial multi-slide modeling with HyperSlides, achieving the most significant performance gains while maintaining a practical training time.

\noindent\textbf{Adaptive Packing.}
Within our batched training framework, we utilize an adaptive packing strategy. One might consider a simpler approach of packing a fixed number of slides (e.g., pairs) to form each hyperslide. However, our investigation reveals that such a fixed-size strategy is suboptimal. As shown in Tab.~\ref{tab:fixed_packing}, fixing the pack size to two slides results in a significantly higher padding ratio (54.9\% vs. 18.5\%). This inefficiency arises because incomplete packs must be padded to a uniform length, leading to wasted computation on uninformative tokens. In contrast, our adaptive packing dynamically fills each hyperslide to its maximum capacity, thereby maximizing token utilization, improving computational efficiency, and yielding superior performance.

\begin{table}[H]
    \small
    \centering
    \setlength{\tabcolsep}{4pt}
    \captionsetup{type=table}
    \caption{Comparison of adaptive and fixed-size packing strategies.}
    \begin{tabular}{lcccc}
        \toprule
        Strategy & Grad. & Sub. & Surv. & Pad. Ratio \\
        \midrule
        Fixed-size (n=2) & 79.48 & 94.54 & 67.78 & 54.9\% \\
        \rowcolor{gray!20}
        Ours (Adaptive) & \textbf{80.19} & \textbf{94.86} & \textbf{68.14} & \textbf{18.5\%} \\
        \bottomrule
    \end{tabular}
    \label{tab:fixed_packing}
\end{table}

\noindent\textbf{Distinction from Mixup-based Augmentation.}
Our HyperSlide methodology is fundamentally distinct from conventional mixup-based data augmentation. The primary distinction lies in the motivation and mechanism. Mixup serves as a regularization technique by creating interpolated training instances and soft labels. In contrast, our goal is to compensate for weak supervision by explicitly modeling inter-slide relationships. We achieve this by constructing clinically meaningful training instances with task-specific macro-labels and a corresponding loss function. This design guides the model to learn from slide ensembles in a clinically relevant manner rather than a purely augmentative one. Furthermore, our approach is length-adaptive, integrating a variable number of slides per pack, unlike typical mixup strategies that operate on fixed pairs. The empirical results in Tab.~\ref{tab:mixup_comparison} corroborate this conceptual difference, showing that a standard mixup approach fails to match the performance of our purpose-built HyperSlide framework.

\begin{table}[t]
    \centering
    \small
    \setlength{\tabcolsep}{4pt}
    \captionsetup{type=table}
    \caption{Performance comparison with slide-level mixup augmentation.}
    \begin{tabular}{lccc}
        \toprule
        Strategy & Grad. & Sub. & Surv. \\
        \midrule
        Slide-level Mixup & 76.25 & 93.23 & N/A \\
        \rowcolor{gray!20}
        Ours (HyperSlide) & \textbf{80.19} & \textbf{94.86} & \textbf{67.50} \\
        \bottomrule
    \end{tabular}
    \label{tab:mixup_comparison}
\end{table}

\subsection{More Discussion about ADS}
\label{sec:app_ads}

While the Attention-driven Downsampler (ADS) module offers benefits in handling redundancy and improving efficiency, its applicability and optimal configuration are subject to certain conditions and data characteristics.

\noindent\textbf{ADS Behavior at Inference}. 
For maintaining interpretability at inference time, the ADS module is configured to preserve per-instance information. This is achieved by disabling the pooling operation along the pack dimension ($k$). However, the learned transformations, including the attention score computation via the shallow MLP, the residual weighting ($u_i=h_i + a_i\,h_i$), and the subsequent linear projection ($v_i = u_i\,W^L$), are still applied to each instance. This allows for the analysis of per-instance attention scores ($a_i$) and transformed features ($v_i$), facilitating post-hoc interpretation of model decisions without reducing the instance count.

\noindent\textbf{Dependence on Data Redundancy}.
The effectiveness of the ADS module is significantly influenced by the inherent redundancy of the input WSI data. For datasets characterized by high tile redundancy, such as TCGA, ADS performs favorably. By reducing the instance count by a factor of $k$, it effectively compresses the representation while discarding redundant or less informative features, leading to computational efficiency and potentially improved signal-to-noise ratio. Conversely, on datasets with intrinsically lower redundancy, like PANDA, applying ADS can be detrimental. In such cases, where a larger proportion of instances may contain clinically relevant information, downsampling can inadvertently discard crucial features, leading to information loss and degraded performance, as shown in Tab.~\ref{tab:ads_panda}. This highlights that the benefits of ADS are most pronounced when applied to data exhibiting substantial spatial redundancy.
\begin{table}[H]
    \small
    \centering
    \captionsetup{type=table}
    \caption{Performance of ADS on PANDA.}
    \begin{tabular}{cccc}
        \toprule
        Strategy & CHIEF & UNI & GIGAP \\
        \midrule
        \rowcolor{gray}w/ ADS   & 76.79   & 78.84   & 77.92   \\
        w/o ADS  & 76.46   & 80.19   & 80.41   \\
        \bottomrule
    \end{tabular}
    \label{tab:ads_panda}
\end{table}

\begin{figure*}[t]
    \centering
    \includegraphics[width=1.\textwidth]{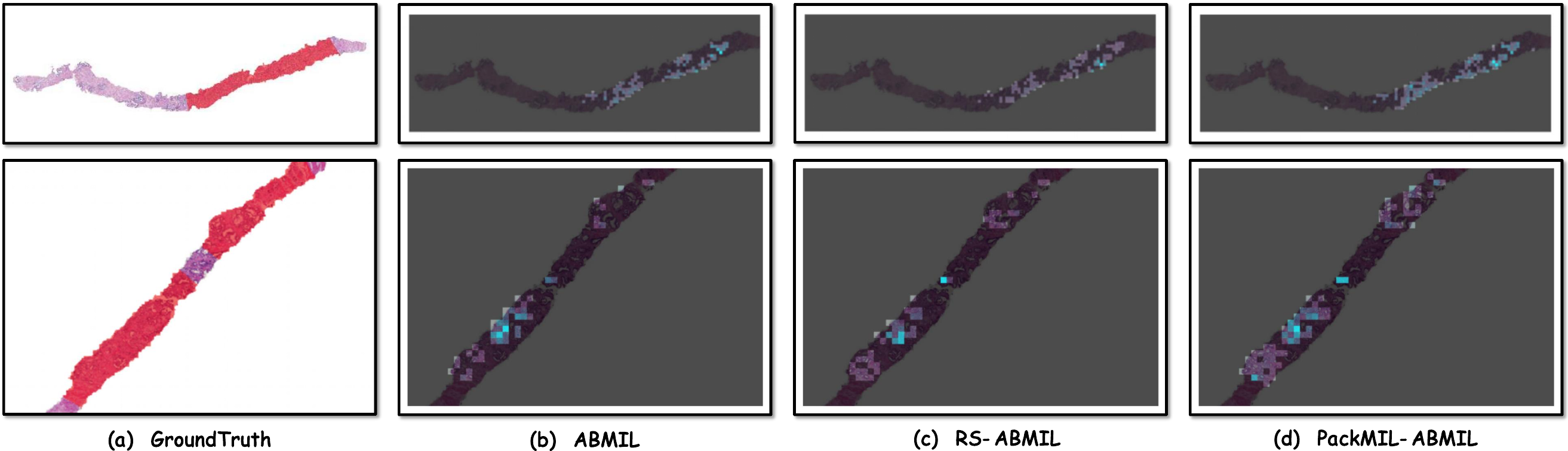}
    \caption{
    Attention visualization on the PANDA dataset~\cite{panda}. The RS strategy, due to its sampling, exhibits limited global attention. With multi-slide supervision via hyperslides and the supplementation of key features, PackMIL demonstrates a more accuracy and comprehensive focus on tumor areas.}
    \label{fig:attention}
    \vspace{-0.3cm}
\end{figure*}
\noindent\textbf{Position of the ADS Module}.
The position of the ADS module is a key impact of its efficacy. As shown in Tab.~\ref{tab:ads_place}, applying the ADS module to the branched feature sets $\mathcal{K}_b$ and $\mathcal{D}_b$ independently (after) yields superior performance over applying it to the combined feature set prior to branching (before). We hypothesize that this is because operating on distinct branches enables the ADS module to learn more specialized attention mechanisms. Such specialization allows the module to better capture the unique characteristics and relevant information within each feature set ($\mathcal{K}_b$ and $\mathcal{D}_b$), which might otherwise be obscured or averaged out in a combined representation.

\begin{table}[H]
    \small
    \centering
    \captionsetup{type=table}
    \caption{Performance of ADS on BRCA.}
    \begin{tabular}{cccc}
        \toprule
        Strategy & CHIEF & UNI & GIGAP \\
        \midrule
        before   & 90.60   & 94.54   & 94.18   \\
        \rowcolor{gray}after  & 92.38   & 94.86   & 94.86   \\
        \bottomrule
    \end{tabular}
    \label{tab:ads_place}
\end{table}

\noindent\textbf{Computational Cost of the ADS Module}.
We evaluate the computational cost and performance impact of incorporating the ADS module. Tab.~\ref{tab:ads_effi} presents a comparison of computational resources and performance metrics for the model trained with and without the ADS component on the BRCA(Subtyping) dataset. The results show that integrating the ADS module requires additional computational resources. Specifically, training time increases from 1 hour to 1.5 hours, memory usage rises from 7GB to 13GB. Importantly, this investment in computational resources is accompanied by a notable improvement in model performance. The model enhanced with the ADS module achieves an AUC of 94.86\%, surpassing the 94.21\% obtained by the baseline model. These findings indicate that while the ADS module introduces additional computational requirements, it effectively contributes to a tangible performance gain.

\begin{table}[H]
    \small
    \centering
    \captionsetup{type=table}
    \caption{Computational Cost of ADS on BRCA.}
    \begin{tabular}{ccccc}
        \toprule
        Method & TTime & Memory & FLOPs & AUC \\
        \midrule
        w/o ADS  & 1h   & 7G    & 49.4G   & 94.21     \\
        \rowcolor{gray}w/ ADS   & 1.5h   & 13G   & 142.1G  & 94.86     \\
        \bottomrule
    \end{tabular}
    \label{tab:ads_effi}
\end{table}

\noindent\textbf{Attention Mechanism}.
The attention mechanism is a critical component of the ADS module. 
In its absence, the downsampling operation would treat all instances uniformly, assigning equal importance to each. 
The core contribution of the attention is to compute instance-specific weights, 
making the feature aggregation process both attention-driven and instance-dependent. This enables the model to selectively focus on and amplify features from the most clinically salient regions. To empirically validate its contribution, we conducted an ablation study by removing the attention component. As shown in Tab.~\ref{tab:ablation_attn}, this resulted in a distinct performance degradation across both subtyping and survival prediction tasks, underscoring the mechanism's importance in learning a meaningful data-driven downsampling policy.

\begin{table}[t]
    \small
    \centering
    \captionsetup{type=table}
    \caption{Performance of attention mechanism for subtyping and survival prediction.}
    \begin{tabular}{ccc}
        \toprule
        Method & Subtyping (AUC) & Survival (C-Index) \\
        \midrule
        w/o Attention & 94.56 & 67.71 \\
        \rowcolor{gray}
        w/ Attention  & 94.86 & 68.14 \\
        \bottomrule
    \end{tabular}
    \label{tab:ablation_attn}
\end{table}

\subsection{Inference Time Comparison}
\begin{table}[t]
\centering
\small
\caption{Inference Efficiency comparison of various MIL methods. Since PackMIL only adds the ADS module during inference and retains the simplest MIL inference pipeline, its inference time is nearly identical to the baseline.}
\label{tab:mil_performance_restructured}
\begin{tabular}{lcc}
\toprule
\textbf{Method} & \textbf{Inference Time} & \textbf{FPS} \\
\midrule
ABMIL & 0.48 ms & 2056 fps \\
DSMIL & 1.04 ms & 960 fps \\
TransMIL & 7.05 ms & 142 fps \\
DTFD & 3.97 ms & 251 fps \\
RRT-MIL & 2.43 ms & 411 fps \\
WIKG & 1.84 ms & 541 fps \\
\midrule
PackMIL(ABMIL) & 0.50 ms & 1984 fps (-4\%) \\
PackMIL(DSMIL) & 1.07 ms & 930 fps (-3\%) \\
PackMIL(VITMIL) & 1.36 ms & 731 fps \\
PackMIL(RRT) & 2.52 ms & 396 fps (-4\%) \\
\bottomrule
\end{tabular}
\end{table}
Tab.~\ref{tab:mil_performance_restructured} presents the inference time with feature input. Since PackMIL only adds the ADS module during inference and retains the simplest MIL inference pipeline, its inference time is nearly identical to the baseline, with less than a 4\% loss in inference speed.

\subsection{Supplementary Confidence Intervals}
We provide the detailed confidence intervals (CI) for our PackMIL-enhanced methods in Tab.~\ref{tab:ci_supplementary}, corresponding to the aggregated results presented in the main manuscript.
\begin{table*}[t]
\small
\centering
\caption{Detailed performance comparison (Mean$_{\pm95\%\text{CI}}$) between RS and PackMIL across all benchmarks.}
\label{tab:ci_supplementary}

\begin{tabular}{ll p{1.4cm}<{\centering} p{1.4cm}<{\centering} p{1.4cm}<{\centering} p{1.4cm}<{\centering} p{1.4cm}<{\centering} p{1.4cm}<{\centering}}
\toprule
\multicolumn{8}{c}{\textbf{(a) Grading on PANDA \& Sub-typing on BRCA}} \\
\midrule
\multicolumn{2}{c}{\multirow{2}{*}{\textbf{Method}}} & \multicolumn{3}{c}{\textbf{Grading (Acc.)}} & \multicolumn{3}{c}{\textbf{Sub-typing (AUC)}} \\
\cmidrule(lr){3-5} \cmidrule(lr){6-8}
\multicolumn{2}{c}{} & CHIEF & UNI & GIGAP & CHIEF & UNI & GIGAP \\
\midrule

 & \textcolor{olive}{+RS} 
& 74.72$_{\pm0.64}$ & 77.91$_{\pm0.58}$ & 78.97$_{\pm0.58}$ 
& 88.72$_{\pm5.39}$ & 93.89$_{\pm3.89}$ & 93.78$_{\pm3.92}$ \\
\rowcolor{gray}
\cellcolor{white}\multirow{-2}{*}{\textbf{ABMIL}} & \textcolor{olive}{+PackMIL}
& 76.46$_{\pm0.61}$ & 80.19$_{\pm0.53}$ & 80.41$_{\pm0.53}$ 
& 92.38$_{\pm4.12}$ & 94.86$_{\pm3.91}$ & 94.86$_{\pm3.68}$ \\
\cmidrule[0.1pt]{1-8}

 & \textcolor{olive}{+RS} 
& 75.00$_{\pm0.65}$ & 78.59$_{\pm0.61}$ & 78.60$_{\pm0.63}$ 
& 91.62$_{\pm4.49}$ & 93.29$_{\pm4.14}$ & 94.04$_{\pm3.97}$ \\
\rowcolor{gray}
\cellcolor{white}\multirow{-2}{*}{\textbf{DSMIL}} & \textcolor{olive}{+PackMIL}
& 75.84$_{\pm0.57}$ & 79.68$_{\pm0.50}$ & 79.10$_{\pm0.55}$ 
& 93.01$_{\pm3.92}$ & 94.62$_{\pm3.72}$ & 94.65$_{\pm3.14}$ \\
\cmidrule[0.1pt]{1-8}

 & \textcolor{olive}{+RS} 
& 73.68$_{\pm0.71}$ & 76.94$_{\pm0.63}$ & 76.15$_{\pm0.67}$ 
& 90.75$_{\pm4.73}$ & 94.07$_{\pm3.89}$ & 93.95$_{\pm3.71}$ \\
\rowcolor{gray}
\cellcolor{white}\multirow{-2}{*}{\textbf{TransMIL}} & \textcolor{olive}{+PackMIL}
& 74.75$_{\pm0.66}$ & 78.87$_{\pm0.57}$ & 78.88$_{\pm0.58}$ 
& 92.31$_{\pm4.23}$ & 94.37$_{\pm3.83}$ & 94.12$_{\pm3.92}$ \\
\cmidrule[0.1pt]{1-8}

 & \textcolor{olive}{+RS} 
& 70.32$_{\pm0.70}$ & 75.04$_{\pm0.60}$ & 75.13$_{\pm0.63}$ 
& 91.55$_{\pm4.67}$ & 94.02$_{\pm3.89}$ & 93.70$_{\pm4.07}$ \\
\rowcolor{gray}
\cellcolor{white}\multirow{-2}{*}{\textbf{RRTMIL}} & \textcolor{olive}{+PackMIL}
& 74.63$_{\pm0.69}$ & 78.46$_{\pm0.59}$ & 78.43$_{\pm0.60}$ 
& 92.43$_{\pm3.92}$ & 94.54$_{\pm3.79}$ & 94.47$_{\pm4.01}$ \\

\bottomrule
\end{tabular}

\vspace{0.4cm}

\begin{tabular}{ll p{1.4cm}<{\centering} p{1.4cm}<{\centering} p{1.4cm}<{\centering} p{1.4cm}<{\centering} p{1.4cm}<{\centering} p{1.4cm}<{\centering}}
\toprule
\multicolumn{8}{c}{\textbf{(b) Survival Analysis on BRCA \& LUAD}} \\
\midrule
\multicolumn{2}{c}{\multirow{2}{*}{\textbf{Method}}} & \multicolumn{3}{c}{\textbf{Survival-BRCA (C-index)}} & \multicolumn{3}{c}{\textbf{Survival-LUAD (C-index)}} \\
\cmidrule(lr){3-5} \cmidrule(lr){6-8}
\multicolumn{2}{c}{} & CHIEF & UNI & GIGAP & CHIEF & UNI & GIGAP \\
\midrule

 & \textcolor{olive}{+RS} 
& 64.02$_{\pm9.2}$ & 65.71$_{\pm9.9}$ & 63.88$_{\pm9.8}$ 
& 61.98$_{\pm8.7}$ & 60.34$_{\pm8.8}$ & 61.01$_{\pm8.6}$ \\
\rowcolor{gray}
\cellcolor{white}\multirow{-2}{*}{\textbf{ABMIL}} & \textcolor{olive}{+PackMIL}
& 68.30$_{\pm8.8}$ & 68.14$_{\pm9.8}$ & 67.04$_{\pm9.1}$ 
& 63.72$_{\pm8.6}$ & 62.60$_{\pm8.5}$ & 61.58$_{\pm8.7}$ \\
\cmidrule[0.1pt]{1-8}

 & \textcolor{olive}{+RS} 
& 65.00$_{\pm9.1}$ & 64.72$_{\pm10.0}$ & 65.69$_{\pm9.8}$ 
& 63.28$_{\pm8.6}$ & 61.53$_{\pm9.1}$ & 61.00$_{\pm8.6}$ \\
\rowcolor{gray}
\cellcolor{white}\multirow{-2}{*}{\textbf{DSMIL}} & \textcolor{olive}{+PackMIL}
& 69.76$_{\pm8.5}$ & 70.00$_{\pm8.6}$ & 68.03$_{\pm9.1}$ 
& 64.10$_{\pm8.6}$ & 62.18$_{\pm8.7}$ & 62.44$_{\pm8.4}$ \\
\cmidrule[0.1pt]{1-8}

 & \textcolor{olive}{+RS} 
& 66.39$_{\pm8.8}$ & 65.14$_{\pm9.8}$ & 65.52$_{\pm9.7}$ 
& 63.53$_{\pm8.8}$ & 62.12$_{\pm8.7}$ & 61.03$_{\pm8.4}$ \\
\rowcolor{gray}
\cellcolor{white}\multirow{-2}{*}{\textbf{TransMIL}} & \textcolor{olive}{+PackMIL}
& 68.08$_{\pm8.6}$ & 68.44$_{\pm8.9}$ & 66.80$_{\pm9.1}$ 
& 64.01$_{\pm9.1}$ & 63.61$_{\pm8.5}$ & 63.04$_{\pm8.3}$ \\
\cmidrule[0.1pt]{1-8}

 & \textcolor{olive}{+RS} 
& 66.11$_{\pm9.3}$ & 64.68$_{\pm10.1}$ & 65.19$_{\pm9.0}$ 
& 62.52$_{\pm8.6}$ & 61.44$_{\pm8.9}$ & 61.35$_{\pm8.5}$ \\
\rowcolor{gray}
\cellcolor{white}\multirow{-2}{*}{\textbf{RRTMIL}} & \textcolor{olive}{+PackMIL}
& 68.15$_{\pm9.1}$ & 68.73$_{\pm9.4}$ & 67.62$_{\pm9.0}$ 
& 64.37$_{\pm8.4}$ & 62.01$_{\pm8.6}$ & 62.79$_{\pm8.2}$ \\

\bottomrule
\end{tabular}
\end{table*}

\section{Qualitative Analysis}
Here, Fig.~\ref{fig:attention} visualizes the attention scores of different MILs on the PANDA. 
We suggest that: 1) Due to the data challenges, traditional non-batched training often struggle to achieve efficient and optimal convergence. As a result, the baseline model (ABMIL) exhibits insufficient discriminability and fails to capture some key pathological details. 2) While the simple random sampling (RS) strategy benefits from improved discriminability through batched training, it suffers from feature loss due to sampling and insufficient supervision, resulting in a lack of global attention. 3) In contrast, PackMIL, leveraging pack-based batched training and multi-slide supervision from hyperslides, demonstrates a more accurate and comprehensive focus on tumor areas.

\section{Additional Methodology Description}
\subsection{Task-specific Hyperslide Loss}
\label{sec:app_task_loss}

\textbf{Grading task.} This task is modeled as single-label classification, we employ Asymmetric Loss (ASLLoss) \cite{ridnik2021asymmetric}. This loss function is chosen to effectively address potential class imbalance and encourage the model to focus on correctly identifying positive classes while being less sensitive to negative misclassifications. Given the predicted probability distribution $\mathbf{p} = [p_1, \dots, p_G]$ for $G$ classes and the corresponding one-hot encoded ground truth labels $\mathbf{y}^\mathrm{hyper}$, the loss is computed as: 
\begin{equation}
\begin{split}
    L_\mathrm{grade}(\mathbf{p}, \mathbf{y}^\mathrm{hyper}) = & -\sum_{i=1}^G \mathbf{y}^\mathrm{hyper}_i (1-p_i)^{\gamma_p} \log(p_i) \\
                                                      & - \sum_{i=1}^G (1-\mathbf{y}^\mathrm{hyper}_i) p_i^{\gamma_n} \log(1-p_i),
\end{split}
\end{equation}
where $\gamma_p \ge 0$ and $\gamma_n \ge 0$ are the focusing parameters for positive and negative samples, respectively.

\begin{table}[H]
    \small
    \centering
    \captionsetup{type=table}
    \caption{Hyperslide loss ablation on PANDA.}
    \begin{tabular}{cccc}
        \toprule
        Strategy & CHIEF & UNI & GIGAP \\
        \midrule
        CE   & 76.46   & 80.16   & 80.09   \\
        \rowcolor{gray}ASL & 76.10   & 80.19   & 80.41   \\
        \bottomrule
    \end{tabular}
    \label{tab:hloss_panda}
\end{table}

\noindent\textbf{Subtyping task.} This task is modeled as multi-label classification with soft targets, we use multi-label Focal Loss \cite{lin2017focal}. This loss helps mitigate the issue of imbalanced subtype frequencies and focuses the model's attention on harder-to-classify samples. Given the predicted probability vector $\mathbf{p} = [p_1, \dots, p_C]^\top$ for $C$ subtypes (typically obtained via Sigmoid activation) and the corresponding soft ground truth label vector $\mathbf{y}^\mathrm{hyper} = [y^\mathrm{hyper}_1, \dots, y^\mathrm{hyper}_C]^\top$, the loss is computed as the sum of binary Focal Loss for each class: 
\begin{equation}
    L_\mathrm{sub}(\mathbf{p}, \mathbf{y}^\mathrm{hyper}) = \sum_{c=1}^C \mathrm{FL}(p_c, y^\mathrm{hyper}_c),
\end{equation} 
where, for class $c$, the binary Focal Loss $\mathrm{FL}(p_c, y^\mathrm{hyper}_c)$ is defined as: 
\begin{equation}
\begin{split}
    \mathrm{FL}(p, y) = &- \alpha y (1-p)^\gamma \log(p) \\
    &- (1-y) (1-\alpha) p^\gamma \log(1-p).
\end{split}
\end{equation}
Here, $\alpha \in [0,1]$ is a weighting factor, and $\gamma \ge 0$ is the focusing parameter.

\begin{table}[H]
    \small
    \vspace{-0.2cm}
    \centering
    \captionsetup{type=table}
    \caption{Hyperslide loss ablation on BRCA.}
    \begin{tabular}{cccc}
        \toprule
        Strategy & CHIEF & UNI & GIGAP \\
        \midrule
        BCE   & 90.89   & 94.20   & 93.91   \\
        \rowcolor{gray}FL & 92.38   & 94.86   & 94.86   \\
        \bottomrule
    \end{tabular}
    \label{tab:hloss_brca}
\end{table}

\noindent\textbf{Survival analysis task.} The model predicts the hazard of event occurrence over discrete time intervals based on hyper-slice features. The training process employs a custom discrete-time Negative Log-Likelihood (NLL) loss function.
Let the follow-up horizon be partitioned into $T$ contiguous, non-overlapping intervals $\{1,\dots ,T\}$.
For individual $i$ the model outputs a hazard sequence
$\mathbf{h}_i=(h_{i,1},\dots ,h_{i,T})$ with
$h_{i,t}=P(T_i=t\mid T_i\ge t,\mathbf x_i)$.
The corresponding discrete survival function is:
\begin{equation}
S_{i,t}\;=\;\prod_{j=1}^{t}(1-h_{i,j}),\qquad t=1,\dots ,T .
\end{equation} 
Denote by
$\delta_i\in\{0,1\}$ the event indicator ($\delta_i=1$ if the event is observed, $0$ if right-censored).
Let $k_i$ be the observed event interval if $\delta_i=1$ and let $k_i^{\mathrm c}$ be the last interval in which the subject is known to be at risk when $\delta_i=0$.
We minimise the following per-sample negative log-likelihood
\begin{equation}
\begin{split}
\mathcal L_i = &
\delta_i\;\underbrace{\bigl[-\log h_{i,k_i} -\log S_{i,k_i}\bigr]}_{\mathcal L_{\text{event},i}}
\;+\; \\
&(1-\delta_i)(1-\alpha)\;\underbrace{\bigl[-\log S_{i,k_i^{\mathrm c}}\bigr]}_{\mathcal L_{\text{cens},i}},
\end{split}
\end{equation} 
where $\alpha\in[0,1]$ down-weights the censored component.
The mini-batch loss is
\begin{equation}
\mathcal L \;=\;\frac1B\sum_{i=1}^{B}\mathcal L_i .
\end{equation} 
Hazards are produced by a sigmoid layer, and survival probabilities are obtained by the cumulative product $S_{i,t}=\prod_{j\le t}(1-h_{i,j})$.
When indices are stored in a 1-based convention, the hazard of the $k$-th interval must be accessed at position $k-1$ of the zero-based tensor, whereas survival $S_{i,k}$ is accessed at position $k$ after prefix-padding with an initial 1.
For censored observations an optional indicator $ \mathbf 1_{k_i=k_i^{\mathrm c}}$ can be applied if the censoring interval exactly coincides with a potential event interval; otherwise every censored instance contributes its survival term.

\subsection{Construction of Isolated Mask}
\label{sec:app_mask}
As described in the method section, all auxiliary masks fall into two functional groups:  
\textit{aggregation-oriented} masks, which constrain feature interaction inside each pack, and  
\textit{classification-oriented} masks, which identify the source bag of every valid token for the downstream classifier.  
We first introduce the shared primitives and then derive both groups.
Let $P_p=\{h_k\}_{k\in\mathcal I_p}$ be the $p$-th pack, padded to length~$L$ and assembled from $B$ bags. 
For positions $j=1,\dots,L$ we define
\begin{align}
(\mathbf{m}_p)_j &= \mathbb{I}\{j\text{ indexes a real feature}\}, & \mathbf{m}_p &\in\{0,1\}^{L}, \label{eq:presence}\\
(\mathbf{b}_p)_j &=
\begin{cases}
\beta_k,& j\text{ contains }h_k,\\
0,& j\text{ is padding},
\end{cases}
& \mathbf{b}_p &\in\{0,\ldots,B\}^{L}, \label{eq:bagid}
\end{align}
where $\beta_k\in\{1,\dots,B\}$ is the global bag index of feature~$h_k$.  
By construction $(\mathbf{m}_p)_j=1$ iff $(p-1)L+j\le M$, with $M$ the total number of tokens before packing.

\noindent\textbf{Aggregation-oriented masks.}
These masks guarantee that feature aggregation never crosses bag boundaries or attends to padding. We construct a binary feature mask and an attention mask for each pack.
Binary feature mask $\mathbf{M}_p \in \{0,1\}^{L \times B}$, indicating which bag each feature belongs to: 
\begin{equation}
\begin{split}
(\mathbf{M}_p)_{j,b} = (\mathbf{m}_p)_j \cdot \mathbb{I}\{(\mathbf{b}_p)_j = b\}, \quad \\ \text{for } j=1,\dots,L, \; b=1,\dots,B.
\label{eq:FeatureMask}
\end{split}
\end{equation} 
An attention mask $\mathbf{A}_p \in \{-\infty, 0\}^{L \times L}$ to enforce intra-bag attention and prevent attention to padding: 
\begin{equation}
\label{eq:AttnMask1}
\mathbf{A}_p \;=\; -\infty\;\Bigl[\mathbf{1}_{L\times L} \;-\; (\mathbf{m}_p\,\mathbf{m}_p^\top)\,\odot\,\mathbf{E}_p\Bigr],
\end{equation}
\begin{equation}
\label{eq:AttnMask2}
(\mathbf{E}_p)_{ij} = \mathbb{I}\{(\mathbf{b}_p)_i = (\mathbf{b}_p)_j\}.
\end{equation}
Here, $\mathbf{1}$ is the all-ones matrix, $\odot$ denotes element-wise multiplication, and $\mathbf{E}_p$ checks if features $i$ and $j$ belong to the same original bag.
Equivalently, $\mathbf{A}_p$ can be visualized in explicit $L\times L$ block-matrix form: 
{\scriptsize
\begin{equation}
\label{eq:AttnMaskBlock}
\begingroup
\setlength{\arraycolsep}{2pt}
\mathbf{A}_p
=
\begin{pmatrix}
\mathbf{0}_{n_1\times n_1} & -\infty\,\mathbf{1}_{n_1\times n_2} & \cdots & -\infty\,\mathbf{1}_{n_1\times n_B} & -\infty\,\mathbf{1}_{n_1\times n_0} \\[6pt]
-\infty\,\mathbf{1}_{n_2\times n_1} & \mathbf{0}_{n_2\times n_2} & \cdots & -\infty\,\mathbf{1}_{n_2\times n_B} & -\infty\,\mathbf{1}_{n_2\times n_0} \\[6pt]
\vdots & \vdots & \ddots & \vdots & \vdots \\[6pt]
-\infty\,\mathbf{1}_{n_B\times n_1} & -\infty\,\mathbf{1}_{n_B\times n_2} & \cdots & \mathbf{0}_{n_B\times n_B} & -\infty\,\mathbf{1}_{n_B\times n_0} \\[6pt]
-\infty\,\mathbf{1}_{n_0\times n_1} & -\infty\,\mathbf{1}_{n_0\times n_2} & \cdots & -\infty\,\mathbf{1}_{n_0\times n_B} & -\infty\,\mathbf{1}_{n_0\times n_0}
\end{pmatrix},
\endgroup
\end{equation}
}
where $$
n_b = \sum_{j=1}^{L} \mathbb{I}\{(\mathbf{b}_p)_j = b\},\quad b=1,\dots,B,
\qquad
n_0 = L - \sum_{b=1}^B n_b
$$ counts tokens from bag $b$ (and $n_0$ counts padding tokens) within pack $p$. $\mathbf{0}_{a\times a}$ is the zero matrix, and $\mathbf{1}_{a\times b}$ is the all-ones matrix.

\noindent\textbf{Classification-oriented masks.}
After aggregation, we must (i) keep only valid tokens and (ii) reveal their bag labels to the classifier.  
Both goals are achieved with
\begin{align}
\mathbf{v}_p &= \mathbf{m}_p, &\quad &\text{(valid-token indicator)}, \label{eq:valid}\\
\mathbf{c}_p &= \mathbf{b}_p, &\quad &\text{(bag-label vector)}. \label{eq:class}
\end{align}
Here $\mathbf{v}_p$ filters out padding positions before the prediction head, whereas $\mathbf{c}_p$ routes each remaining token to the correct bag-level logit.

Aggregation-oriented masks act inside the encoder to enforce intra-bag interactions, while classification-oriented masks operate at the output stage to attach each valid token to its original bag.  
Together they preserve bag integrity and enable efficient batched processing without introducing extra parameters.

\subsection{Dynamic Pack Length Adaptation}
While the pack operation utilizes a fixed length $L$ for efficient batch processing, the actual number of instances sampled from each bag ($|\tilde{\mathcal{K}}_b|$ and $|\tilde{\mathcal{D}}_b|$) can vary significantly due to stochastic sampling and the diverse sizes of original WSIs.
To accommodate this inherent variability, particularly when a mini-batch contains bags that yield a large number of sampled instances, we incorporate a dynamic pack length adaptation mechanism. Before packing the instances for a given branch (main or residual) within a mini-batch, we assess the maximum sampled sequence length from any single bag in that batch. Specifically, we check if $\max_b |\tilde{\mathcal{K}}_b|$ (for the main branch) or $\max_b |\tilde{\mathcal{D}}_b|$ (for the residual branch) exceeds the current pack length $L$.
If this condition is met, the pack length for that specific branch and mini-batch is dynamically doubled to $2L$. This dynamic doubling occurs at most once per branch per mini-batch processing step, effectively setting an upper bound of $2L$ on the pack length. This adaptation ensures that sampled instances from bags with particularly large retained or discarded sets are less likely to be fragmented across numerous packs, leading to more efficient packing and potentially better representation within packs for such cases.
\begin{table}[H]
    \small
    \centering
    \captionsetup{type=table}
    \caption{Ablation on Dynamic Pack Length.}
    \begin{tabular}{cccc}
        \toprule
        Strategy & Grad. & Sub. & Surv. \\
        \midrule
        Fixed Length   & 79.69   & 94.42   & 67.72   \\
        \rowcolor{gray}Dynamic Length & 80.19   & 94.86   & 68.14   \\
        \bottomrule
    \end{tabular}
    \label{tab:ablation_dylength}
\end{table}

\subsection{Pseudo-code}

Algorithm~\ref{algo:ads} outlines the PyTorch-style pseudocode for the ADS module. Subsequently, Algorithm~\ref{algo:packing} details the logic of our packing strategy.

\begin{algorithm*}[t]
  \setstretch{0.85}
  \PyComment{x: input instance features, shape $[B, N, D]$} \\
  \PyComment{k: downsampling factor} \\
  \PyComment{W\_L, W\_P: learnable linear projections} \\
  \PyComment{attn: attention score} \\
  ~\\
  \PyComment{1. Compute attention-driven residual features} \\
  \PyCode{alpha = attn(x).softmax(dim=1)} \\
  \PyCode{x = x + x * alpha} \\
  ~\\
  \PyComment{2. Linear transformation} \\
  \PyCode{x = x @ W\_L} \\
  ~\\
  \PyComment{3. Prepare for Grouping (Training vs Inference)} \\
  \PyCode{if training:} \\
  \Indp
      \PyComment{Stochastic mode: Shuffle for diversity} \\
      \PyCode{x = shuffle(x, dim=1)} \\
  \Indm
  \PyCode{else:} \\
  \Indp
      \PyComment{Deterministic mode: Keep order for interpretability} \\
      \PyCode{pool\_mode = None}  \\
  \Indm
  ~\\
  \PyComment{4. Instance Unshuffle and Pooling} \\
  \PyCode{B, N, D = x.shape} \\
  \PyCode{groups = x.view(B, N // k, k, D)} \PyComment{Reshape into groups of size k} \\
  ~\\
  \PyCode{if mode == 'max':} \\
  \Indp
      \PyCode{x\_pooled = groups.max(dim=2).values} \\
  \Indm
  \PyCode{elif mode == 'random':} \\
  \Indp
      \PyCode{rand\_idx = torch.randint(0, k, (B, N // k))} \\
      \PyCode{x\_pooled = groups.gather(2, rand\_idx)} \\
  \Indm
  \PyCode{else:} \\
  \Indp
      \PyComment{Inference} \\
      \PyCode{x\_pooled = x} \\
  \Indm
  ~\\
  \PyComment{5. Final projection} \\
  \PyCode{x\_out = x\_pooled @ W\_P} \\
  
  \caption{PyTorch-style pseudocode for ADS (Training vs. Inference)}
  \label{algo:ads}
\end{algorithm*}

\begin{algorithm*}[t]
  \setstretch{0.85}
  \PyComment{batch: list of WSI feature sets, each shape $[N_i, D]$} \\
  \PyComment{L: target fixed pack length} \\
  \PyComment{r: keep ratio for main branch} \\
  \PyComment{ADS: Attention-driven Downsampler module} \\
  ~\\
  \PyCode{buffer\_main, buffer\_res = [], []} \\
  ~\\
  \PyComment{1. Process each bag: Sample and Downsample} \\
  \PyCode{for x in batch:} \\
  \Indp
      \PyCode{N = x.shape[0]} \\
      \PyComment{Stochastic instance-level sampling} \\
      \PyCode{perm = torch.randperm(N)} \\
      \PyCode{num\_keep = int(N * r)} \\
      \PyCode{idx\_keep, idx\_disc = perm[:num\_keep], perm[num\_keep:]} \\
      ~\\
      \PyComment{Apply ADS to obtain compact representations (ensure min/max length constraint)} \\
      \PyCode{x\_keep = ADS(x[idx\_keep])} \PyComment{Shape: $[n_{keep}, D]$} \\
      \PyCode{x\_disc = ADS(x[idx\_disc])} \PyComment{Shape: $[n_{disc}, D]$} \\
      ~\\
      \PyCode{buffer\_main.append(x\_keep)} \\
      \PyCode{buffer\_res.append(x\_disc)} \\
  \Indm
  ~\\
  \PyComment{2. Sequential Packing Operation (Next-Fit Strategy)} \\
  \PyCode{def run\_packing(feature\_list, L):} \\
  \Indp
      \PyCode{packs = []} \\
      \PyCode{current\_pack = []} \\
      \PyCode{current\_len = 0} \\
      ~\\
      \PyCode{for x in feature\_list:} \PyComment{x: current bag features, Shape $[n, D]$} \\
      \Indp
          \PyCode{n = x.shape[0]} \\
          \PyComment{Check if current bag fits in current pack} \\
          \PyCode{if current\_len + n > L:} \\
          \Indp
              \PyComment{Pack is full: Pad remaining slots with zeros} \\
              \PyCode{pad\_size = L - current\_len} \\
              \PyCode{finished\_pack = concat(current\_pack + [zeros(pad\_size, D)])} \PyComment{Shape: $[L, D]$} \\
              \PyCode{packs.append(finished\_pack)} \\
              ~\\
              \PyComment{Start new pack with current bag} \\
              \PyCode{current\_pack = [x]} \\
              \PyCode{current\_len = n} \\
          \Indm
          \PyCode{else:} \\
          \Indp
              \PyComment{Append bag to current pack without splitting} \\
              \PyCode{current\_pack.append(x)} \\
              \PyCode{current\_len = current\_len + n} \\
          \Indm
      \Indm
      ~\\
      \PyComment{Handle the last remaining pack} \\
      \PyCode{if current\_len > 0:} \\
      \Indp
          \PyCode{pad\_size = L - current\_len} \\
          \PyCode{last\_pack = concat(current\_pack + [zeros(pad\_size, D)])} \PyComment{Shape: $[L, D]$} \\
          \PyCode{packs.append(last\_pack)} \\
      \Indm
      ~\\
      \PyCode{return stack(packs)} \PyComment{Final Output Shape: $[B', L, D]$} \\
  \Indm
  ~\\
  \PyCode{P\_main = run\_packing(buffer\_main, L)} \\
  \PyCode{P\_res = run\_packing(buffer\_res, L)} \\
  
  \caption{PyTorch-style pseudocode for Stochastic Sampling and Sequential Packing}
  \label{algo:packing}
\end{algorithm*}

\section{Additional Related Works}
\label{sec:app_rw}
\subsection{Pack-based Batched Training}
Training on variable-length sequences has traditionally relied on padding shorter sequences and applying masks to ignore padded tokens, ensuring uniform batch shapes at the cost of wasted computation~\cite{krell2021efficient}.
To reduce this overhead, dynamic batching (length-based bucketing) groups sequences of similar lengths per batch, greatly minimizing padding requirements~\cite{zelasko2025emmett}.
Modern transformer architectures further exploit attention masks to handle padding, and recent work goes beyond simple padding by packing multiple sequences into one longer sequence with special separators and adjusted position indices~\cite{krell2021efficient}.
Such sequence packing techniques, originally used in large-scale NLP pre-training, can double throughput by eliminating pad tokens~\cite{kosec2021packing} while maintaining model fidelity via careful masking to prevent cross-sequence attention.
For example, packing algorithms in BERT pre-training combine several short sentences into a single 512-token input, yielding ~2× speedups with negligible accuracy loss~\cite{kosec2021packing}.
In computer vision, analogous ideas enable variable-resolution training. NaViT avoids fixed-size resizing by treating images as sequences of patches and packing arbitrary-resolution inputs, improving efficiency in large-scale image–text pre-training without sacrificing performance~\cite{dehghani2023patch}.
Other works dynamically reduce sequence length during processing, such as Token Merging (ToMe) merges redundant tokens in ViTs to halve the token count on the fly, boosting throughput ~2× for large models with minimal accuracy drop~\cite{bolya2022token}.
RNN-based systems commonly use packed sequences to skip computation on padded timesteps, and Transformer-based LLMs and vision models use padding masks or adaptive token pruning to similar effect.
In CPath, where WSI yields a bag of thousands of instances, efficient batching is critical. However, the data characteristics in CPath render the direct application of the aforementioned strategies non-trivial. Approaches focused on packing short sequences provide limited benefits, while sampling long sequences risks significant information loss. Effectively adapting these efficient training paradigms to CPath is thus a key challenge. Our proposed pack-based framework addresses this by incorporating a residual branch to more effectively packing these variable-length long sequences, aiming to mitigate these limitations.

\subsection{More about Batchsize in Computational Pathology}
As elaborated in the \textit{Related Work} section, current slide-level MIL methods often train with batchsize of 1.
Conversely, when explicit patch-level annotations are available for segmentation or detection tasks, researchers commonly employ moderate to large batch sizes by independently processing uniformly-sized patches extracted from WSIs, enabling stable training and efficient convergence~\cite{ciga2021overcoming,graham2019hover}.
These models independently process uniformly-sized patches extracted from WSIs, leveraging moderate to large batch sizes to facilitate stable training and efficient convergence~\cite{ciga2021overcoming,graham2019hover}.
Conversely, when explicit patch-level annotations are available for segmentation or detection tasks, researchers typically form batches at the patch level. 
They independently process uniformly-sized patches from WSIs using moderate to large batchsizes, facilitating stable training and efficient convergence~\cite{ciga2021overcoming,graham2019hover}. 
Such patch-centric batching, however, presents a discrepancy with the holistic slide assessment in clinical workflows.

\section{Conclusion}

In CPath, gigapixel WSIs exhibit extremely long sequences, significant length variations, high redundancy, and limited supervision. 
Existing methods typically address only individual aspects of these challenges, lacking systematic exploration.
Our work reveals that these challenges lead to training inefficiency, instability, and high redundancy on both large-scale and conventional datasets. 
To comprehensively tackle these issues, we propose the pack-based MIL framework. 
It enables batch training while preserving data heterogeneity, enhancing both training efficiency and quality by a large margin. 
We incorporates a residual branch that utilizes \textit{hyperslides} to supplements the limited supervision while mitigating feature loss from packing.
Moreover, a attention-driven downsampler is integrated to compress feature redundancy within both branches. 
We also systematically evaluated a simple random sampling training strategy, which demonstrated considerable improvements on the PANDA. 
With extensive experiments, we summarize practical guidelines for batched CPath training and highlight the significant potential of focusing on data challenges in the era of FM.

\section{Limitation \& Broader Impacts}
This work revisiting data challenges in computational pathology and, by considering these challenges, proposes a pack-based MIL training framework. However, the primary limitation of this method is the significant challenge in implementing batched training of complex MIL models based on packs. While we have implemented with commonly used models such as ABMIL, TransMIL, and DSMIL, constructing the required masks for some more complex model structures remains challenging. Furthermore, the current hyperslide training is highly specific to downstream tasks. Designing a more general training objective is a key focus of our future work. Beyond these limitations, this work holds significant potential to advance key healthcare tasks such as cancer diagnosis and prognosis. The significant performance improvements demonstrated in this work, particularly when leveraging foundation model features, hold potential to benefit and inspire the development of more accurate state-of-the-art algorithms in the clinical scenario.

\section{Data Availability Statement}

The PANDA dataset (CC-BY-4.0) is available at \url{https://panda.grand-challenge.org/}. 

All TCGA datasets can be found at \url{https://portal.gdc.cancer.gov/}.

The CAMELYON dataset is available at~\url{https://camelyon17.grand-challenge.org/}.

The Diabetic Retinopathy Grading dataset is available at~\url{https://github.com/nkicsl/DDR-dataset}.

\end{document}